\documentclass{article} 
\usepackage{iclr2024_conference,times}

\usepackage{amsmath,amsfonts,bm}

\def\eqref#1{equation~\ref{#1}}

\def\1{\bm{1}}

\DeclareMathAlphabet{\mathsfit}{\encodingdefault}{\sfdefault}{m}{sl}
\SetMathAlphabet{\mathsfit}{bold}{\encodingdefault}{\sfdefault}{bx}{n}

\usepackage{url}

\usepackage[pagebackref,breaklinks,colorlinks]{hyperref}

\usepackage[utf8]{inputenc} 
\usepackage[T1]{fontenc}    

\usepackage{booktabs}
\usepackage{graphicx}
\usepackage{tablefootnote}
\usepackage{bbm}

\usepackage{cite}
\usepackage{amsmath,amssymb,amsfonts}
\usepackage{textcomp}
\usepackage{subcaption}
\usepackage{tabularx}
\usepackage{booktabs}       
\usepackage[normalem]{ulem}
\useunder{\uline}{\ul}{}
\usepackage{bm}
\usepackage{color}
\usepackage{xcolor}
\usepackage{algorithm}
\usepackage{algorithmicx}
\usepackage{algpseudocode}
\usepackage{multirow}
\usepackage{xspace}

\usepackage[utf8]{inputenc}
\usepackage[T1]{fontenc}
\usepackage[margin=1in]{geometry}
\usepackage{times}

\usepackage{algpseudocode}
\usepackage{enumitem}
\usepackage{lipsum}
\usepackage{hyperref}

\usepackage{booktabs}
\usepackage{colortbl}
\usepackage{xcolor}
\usepackage{multirow}

\definecolor{lightgray}{gray}{0.9}

\hypersetup{
    colorlinks=true,
    linkcolor=blue,
    filecolor=magenta,      
    urlcolor=cyan,
    citecolor=blue,
}

\newcommand{\agentset}[1]{\mathcal{#1}}
\newcommand{\symboletongyi}{\raisebox{0pt}{~\includegraphics[scale=0.05]{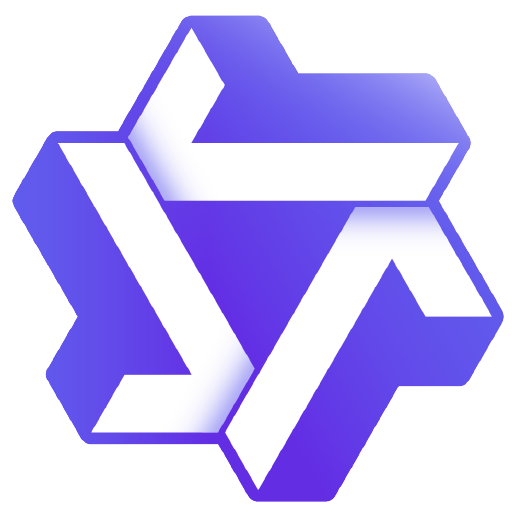}}~}

\usepackage{threeparttable}
\usepackage{newtxtext}
\usepackage[capitalize]{cleveref}
\crefname{section}{Sec.}{Secs.}
\Crefname{section}{Section}{Sections}
\Crefname{table}{Table}{Tables}
\crefname{table}{Tab.}{Tabs.}

\definecolor{demphcolor}{RGB}{144, 144, 144}
\definecolor{mygray}{gray}{0.4}
\definecolor{lightgray}{rgb}{0.9, 0.9, 0.9}
\definecolor{deepgreen}{RGB}{0,100,0}

\hypersetup{%
  citecolor=teal
}
\hypersetup{linkcolor = black}
\newcommand{\modelname}{GUI-Owl\xspace}

\title{Mobile-Agent-v3: Fundamental Agents for GUI Automation}

\usepackage[symbol]{footmisc}

\iclrfinalarxiv 
\begin{document}

\author{
}

\maketitle

\begin{center}
   \centering
   \vspace{-16mm}
   \textbf{Jiabo Ye\footnote{Equal contribution}\qquad Xi Zhang\footnotemark[1] \qquad Haiyang Xu\footnotemark[1]\footnotemark[2] \qquad Haowei Liu\qquad Junyang Wang \qquad Zhaoqing Zhu \qquad Ziwei Zheng \qquad Feiyu Gao \qquad Junjie Cao \qquad Zhengxi Lu \\ \qquad Jitong Liao \qquad Qi Zheng \qquad Fei Huang \qquad Jingren Zhou \qquad Ming Yan\footnote{Corresponding author and project leader}} \\
    {Tongyi Lab\symboletongyi, Alibaba Group}\\
    {\tt\small \{shuofeng.xhy, ym119608\}@alibaba-inc.com} 
   
   \url{https://github.com/X-PLUG/MobileAgent}
\end{center}

\begin{abstract}

This paper introduces GUI-Owl, a foundational GUI agent model that achieves new state-of-the-art performance among open-source end-to-end models across ten GUI benchmarks spanning both desktop and mobile environments, covering grounding, question answering, planning, decision-making, and general procedural knowledge in GUI automation scenarios. Notably, GUI-Owl-7B achieves a score of 66.4 on the AndroidWorld benchmark and 29.4 on the OSWorld-Verified benchmark. Building on this model, we propose a general-purpose GUI agent framework, Mobile-Agent-v3, which further enhances GUI-Owl’s performance (73.3 on AndroidWorld and 37.7 on OSWorld-Verified), achieving a new state-of-the-art among GUI agent frameworks based on open-source models. GUI-Owl incorporates several key innovations: 
1) \textbf{Large-scale Environment Infrastructure}: We introduce a cloud-based virtual environment infrastructure spanning different operating systems (including Android, Ubuntu, macOS, and Windows). 
This underpins our Self-Evolving GUI Trajectory Production framework, which generates high-quality interaction data through sophisticated query generation and correctness judgment. The framework leverages GUI-Owl's capabilities to continuously refine trajectories, creating a self-reinforcing improvement cycle. It supports multiple downstream data pipelines, enabling robust data collection while reducing manual annotation needs.
%
2) \textbf{Diverse Foundational Agents Capability Construction}: by incorporating foundational UI data—such as grounding, planning, and action semantic recognition—alongside diverse reasoning and reflecting patterns, GUI-Owl not only supports end-to-end decision making but can also serve as a specialized module integrated into multi-agent frameworks; 3) \textbf{Scalable Environment RL}: we also develop a scalable reinforcement learning framework that enables fully asynchronous training and better aligns the model’s decision with real-world usage. In addition, we introduce Trajectory-aware Relative Policy Optimization (TRPO) for online environment RL, which achieves 34.9 on the OSWorld-Verified benchmark. GUI-Owl and Mobile-Agent-v3 are open-sourced at: \url{https://github.com/X-PLUG/MobileAgent}.

\end{abstract}

\begin{figure}[h]
    \centering
    \includegraphics[width=\textwidth]{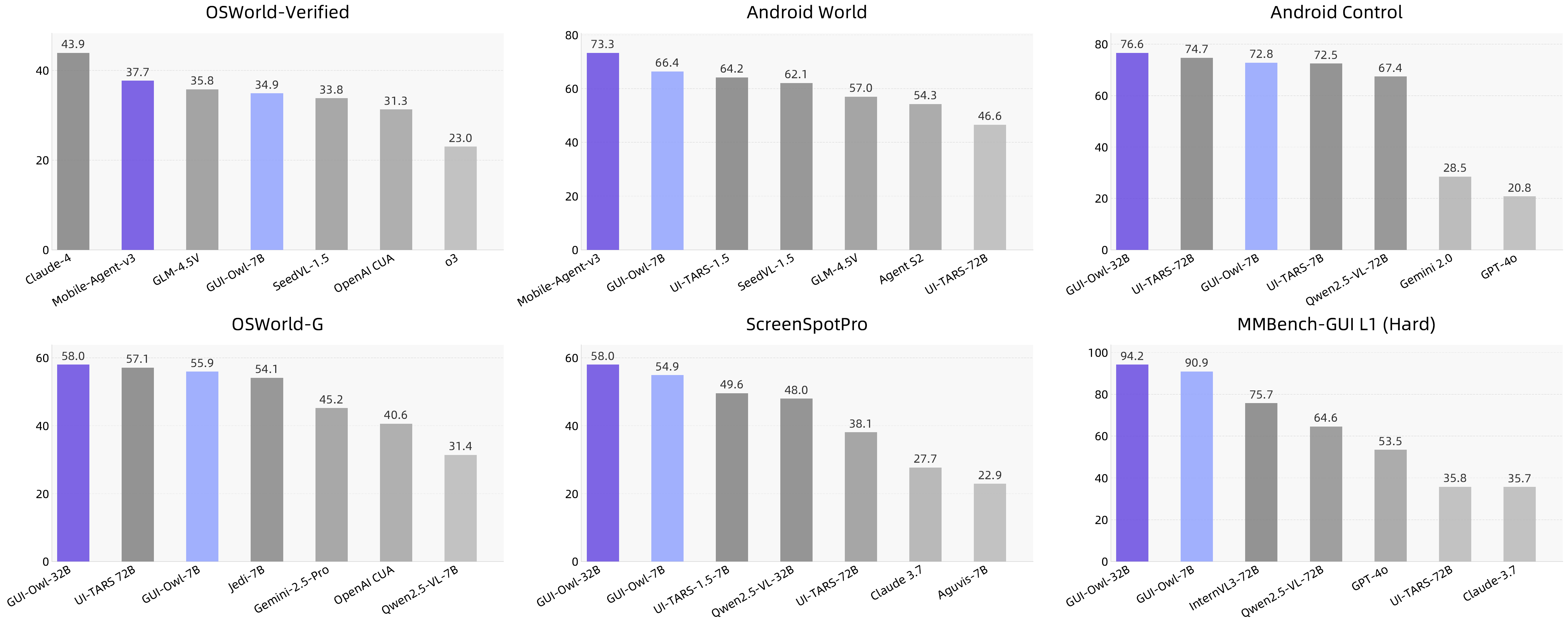}
    \caption{Performance overview on mainstream GUI-automation benchmarks.}
    \label{fig:score}
\end{figure}

\section{Introduction}
Graphical User Interfaces (GUIs) Agents~\citep{hu2024agents,zhang2024large,nguyen2024gui,wang2024gui,gao2024generalist,wang2024mobile} is designed to automate daily and professional tasks based on human instructions across various device environments, thereby enhancing production efficiency. With the rapid advancement of multimodal large models and reasoning technologies, vision-based GUI agents have demonstrated strong task execution capabilities across various device environments, including PCs, mobile devices, and web platforms.

The existing methods can be broadly divided into two categories. The first category builds agent frameworks based on closed-source models~\citep{yang2025gta1,xie2025scalingcomputerusegroundinguser,Agent-S2,song2025coact,wang2024mobile2}, however, these approaches struggle to handle unfamiliar tasks and adapt to dynamic environments. The second category focuses mainly on end-to-end model performance~\citep{qin2025ui,wang2025opencuaopenfoundationscomputeruse}, but such methods often fail to follow instructions faithfully and lack compatibility with diverse agent frameworks, significantly limiting their practical utility. The GUI agents require this foundational model to have the following capabilities: 1)  The strong UI perception capabilities (such as for Mobile, PC, and Web); 2) The planning, reflection, and reasoning in various dynamic environments; 3) The flexibility to integrate with various multi-agent frameworks.

\begin{figure}[t]
    \centering
    \includegraphics[width=0.8\textwidth]{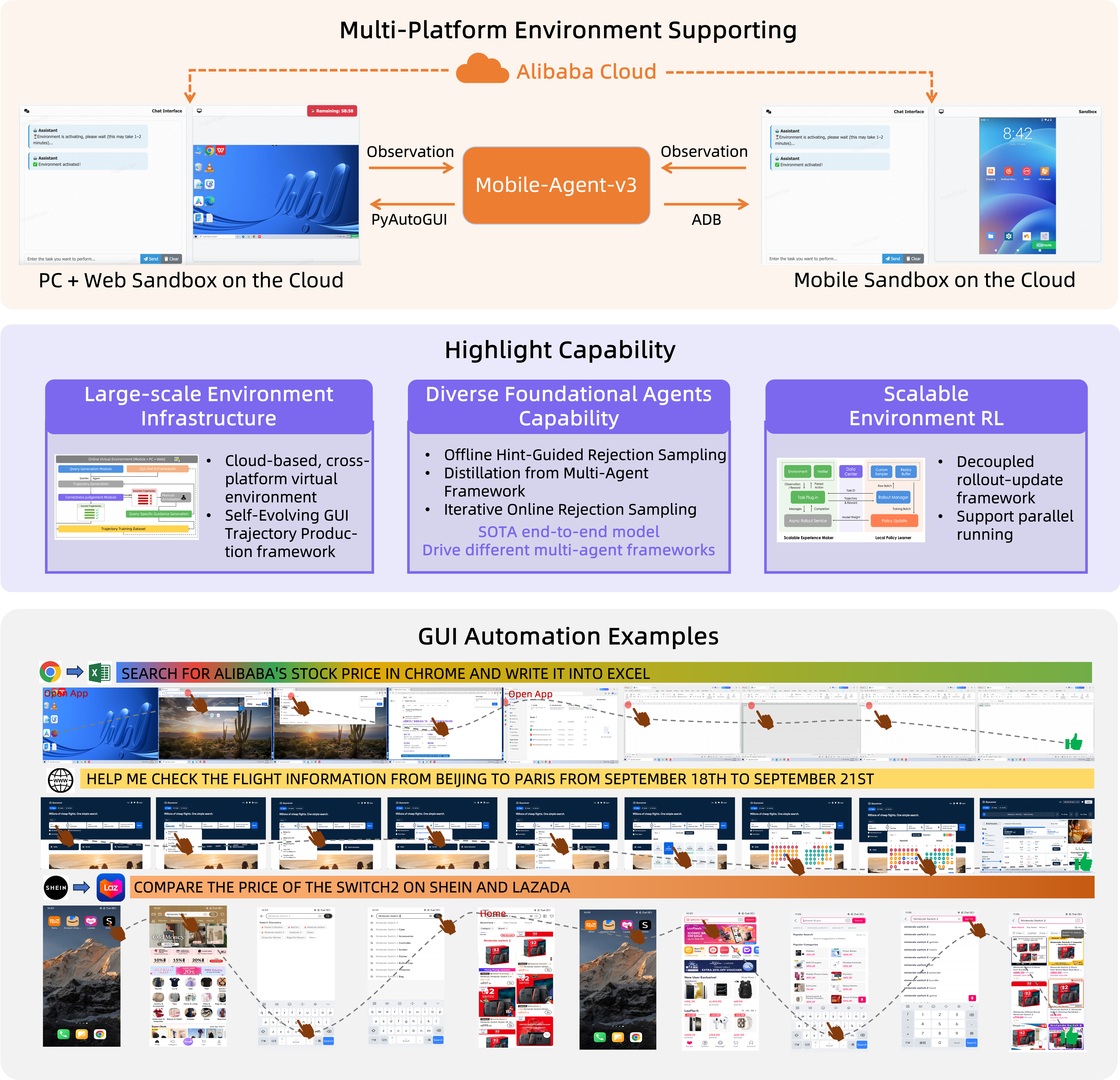}
    \caption{Overview of our Mobile-Agent-v3. We illustrate our multi-platform environment supporting, our core capability, and some GUI automation examples generated by Mobile-Agent-v3.}
    \label{fig:2}
\end{figure}

In this paper, we propose GUI-Owl, a native end-to-end multimodal agent designed as a foundational model for GUI automation. Built upon Qwen2.5-VL and extensively post-trained on large-scale, diverse GUI interaction data, GUI-Owl unifies perception, grounding, reasoning, planning, and action execution within a single policy network. The model achieves robust cross-platform interaction, handling multi-turn decision making with explicit intermediate reasoning, and supports both autonomous operation and role-specific deployment in multi-agent systems. To further enhance its adaptability, we develop specialized datasets for core foundational tasks—including UI grounding, task planning, and action semantics—and employ a scalable reinforcement learning framework to align GUI-Owl’s decision policy with real-world task success. Beyond single-agent deployment, GUI-Owl can be instantiated as different specialized agents within a multi-agent framework Mobile-Agent-v3 where multiple role agents coordinate and share partial observations and reasoning traces to tackle complex, long-horizon automation workflows.

\paragraph{Large-scale Environment Infrastructure.}
To train our GUI agent, we developed a comprehensive large-scale environment infrastructure for GUI interaction data collection. This infrastructure leverages cloud-based technologies, including cloud phones and cloud computers on Alibaba Cloud~\citep{cloud2018introducing}, spanning mobile, PC, and web platforms. It creates dynamic virtual environments that enable diverse and realistic interaction scenarios across various operating systems and devices.
Central to this infrastructure is our Self-Evolving GUI Trajectory Production pipeline. This innovative system collects high-quality trajectory data through a sophisticated process involving: high-quality query generation that mimics real-world user interactions, model roll-outs where GUI-Owl and Mobile-Agent-v3 interacts with virtual environments, rigorous correctness judgment to ensure data quality, and query-specific guidance generation for challenging scenarios.
This self-evolving pipeline creates a continuous improvement cycle, enhancing both our dataset quality and the GUI-Owl model's capabilities over time. 
By combining cloud technology with multi-platform environments, our infrastructure enables efficient, scalable model development while reducing manual annotation needs.


%
%
\paragraph{Diverse Foundational Agents Capability Construction.} Based on the generated trajectories, we introduce multiple downstream data construction pipelines to enhance the agent’s fundamental UI capabilities, including: (i) a grounding pipeline that covers both UI element localization—based on functional, appearance, and layout instructions—and fine-grained word/character grounding; (ii) a task planning pipeline that distills procedural knowledge from historical successful trajectories and large-scale pretrained LLMs to handle long-horizon, multi-application tasks; and (iii) an action semantics pipeline that captures the relationship between actions and resulting state transitions through before/after UI observations. Furthermore, we synthesize reasoning and reflecting data using offline hint-guided rejection sampling, distillation from a multi-agent framework, and iterative online rejection sampling. This supervision enables the agent to perform independent reasoning and to engage in complex, long-horizon collaborative reasoning within the Mobile-Agent-v3 framework, adapting its reasoning style to the specific role it assumes. 

\paragraph{Scalable Environment RL.} We also develop a scalable training framework grounded in a unified interface for multi-task training that standardizes interactions for both single-turn reasoning and multi-turn agentic tasks, and we decouple experience generation from policy updates to provide fine-grained control over policy adherence. This design supports fully asynchronous training and better aligns the model’s decision-making with real-world usage. We further introduce Trajectory-aware Relative Policy Optimization (TRPO) to address training with long, variable-length action sequences in online-environment RL. TRPO uses trajectory-level rewards to compute step-level advantages and employs a replay buffer to improve the stability of reinforcement learning.

We evaluate \modelname across a wide range of benchmarks that comprehensively measure native agent capability on GUI automation including grounding, single-step decision, general question answering and evaluate on online environment. \modelname-7B outperforms all state-of-the-art models of comparable size. In particular, \modelname-7B achieves scores of 34.9 on OSWorld-Verified and 66.4 on AndroidWorld. Moreover, \modelname-32B demonstrates outstanding performance, surpassing even proprietary models. On MMBench-GUI and AndroidControl, GUI-Owl-32B outperforms all models, including GPT-4o and Claude 3.7. In grounding capability evaluations, \modelname-32B surpasses all models of the same size and achieves competitive performance compared with proprietary models. When combined with Mobile-Agent-v3, it achieves scores of 37.7 on OSWorld and 73.3 on AndroidWorld, which clearly demonstrates its capability as a fundamental agent for GUI automation.

\section{GUI-Owl}
\modelname is an end-to-end multimodal model that unifies capabilities such as perception, planning, decision-making, and grounding within GUI scenarios. By leveraging extensive and diverse datasets for post-training based on Qwen2.5-VL, GUI-Owl is able to interact with graphical user interfaces on Mobile, PC, and Web platforms. We further apply reinforcement learning to GUI-Owl to align its capabilities with diverse downstream requirements. This alignment enables the model not only to autonomously perform multi-turn GUI interaction tasks, but also to generalize to specific applications such as question answering, captioning, planning, and grounding. Moreover, GUI-Owl can assume various roles within a multi-agent framework, in which individual agents fulfill their respective responsibilities, coordinate their actions, and collaboratively accomplish more complex tasks.

\subsection{End-to-end GUI interactions}
\begin{figure*}[!ht]
    \centering
    \includegraphics[width=\textwidth]{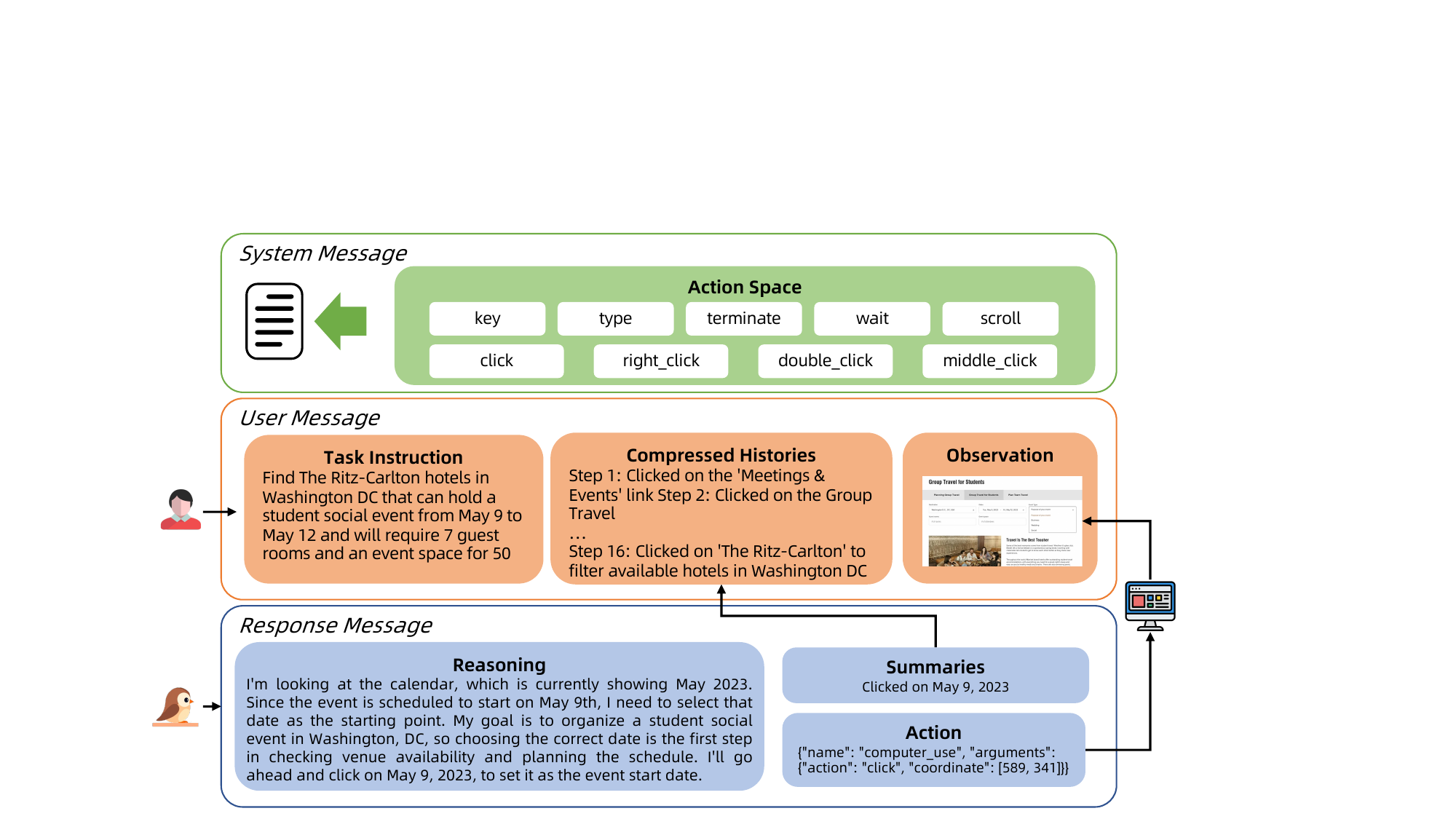}
    \caption{Illustration of the interaction flow of \modelname. The system message defines the available action space, the user message contains the task instruction, compressed histories, and current observation, while the response message includes the agent's reasoning, action summaries, and the final action output.}
    \label{fig:gui_e2e}
\end{figure*}
We model the interaction process between \textbf{GUI-Owl} and the device, as well as the completion of the specified task, as a \textit{multi-turn decision-making process}. Given the available action space $A = \{ a^1, a^2, \dots, a^{|A|} \}$ of the environment, the current environment observation $S_t\in \mathcal{S}$, which can be a screenshot in common, and the history of past operations $H_t = \{(S_1, a_1), (S_2, a_2), \dots, (S_{t-1}, a_{t-1})\}$, the model selects an action from the action space $A$ and executes it in the environment to obtain the next time-step's observation $S_{t+1}$.

Formally, at each time step $t$, $ a_t \sim \pi( \cdot \mid S_t, H_t ), $ here, $\pi$ denotes policy model (\modelname), which maps the current observation and historical operations to a probability distribution over the action space $A$. 

We present the interaction flow of \modelname in \Cref{fig:gui_e2e}. In practice, we support flexible prompts to organize the action space into system messages. By default, we adopt the Qwen function calling format. Detail action space definition is presented in \Cref{tab:space_e2e_mobile} and \Cref{tab:space_e2e_desktop}. For user messages, we sequentially provide the original task, historical information, and observations. To save GPU memory and improve inference speed, we typically retain only the most recent 1 to 3 images. Notably, requiring a robust reasoning process before the actual output of an action decision can enhance the model's ability to adapt to complex tasks and situations. Therefore, we require the model to first "reasoning" before making a decision, and then execute the action based on this reasoning content. However, since lengthy thoughts over multiple turn interactions may cause the conversation history to become excessively long, we additionally require the model to output a "conclusion" summarizing the key information of the current step. Finally, only the conclusion is stored in the historical context.

The actions output by the model are translated into actual device operation commands (for example, we use ADB commands for Android devices, and pyautogui code for desktop operations). Meanwhile, the latest screenshot of the device's display is further captured and used as the observation of the environment.

\subsection{Foundational Agents Capability}

\modelname can function not only as a native agent capable of independently interacting with GUIs, but also provides a variety of foundational capabilities to support downstream standalone calls or integration into a multi-agent framework. To this end, we collect and construct datasets for various capabilities such as grounding, caption and planning. These datasets are mixed with general instruction data during training, and we found that the model also possesses zero-shot GUI question-answering capability as well as general instruction-following abilities for unseen tasks.

\subsubsection{Self-Evolving Trajectory Production Framework}

\begin{figure*}[htpb]
    \centering
    \includegraphics[width=0.8\textwidth]{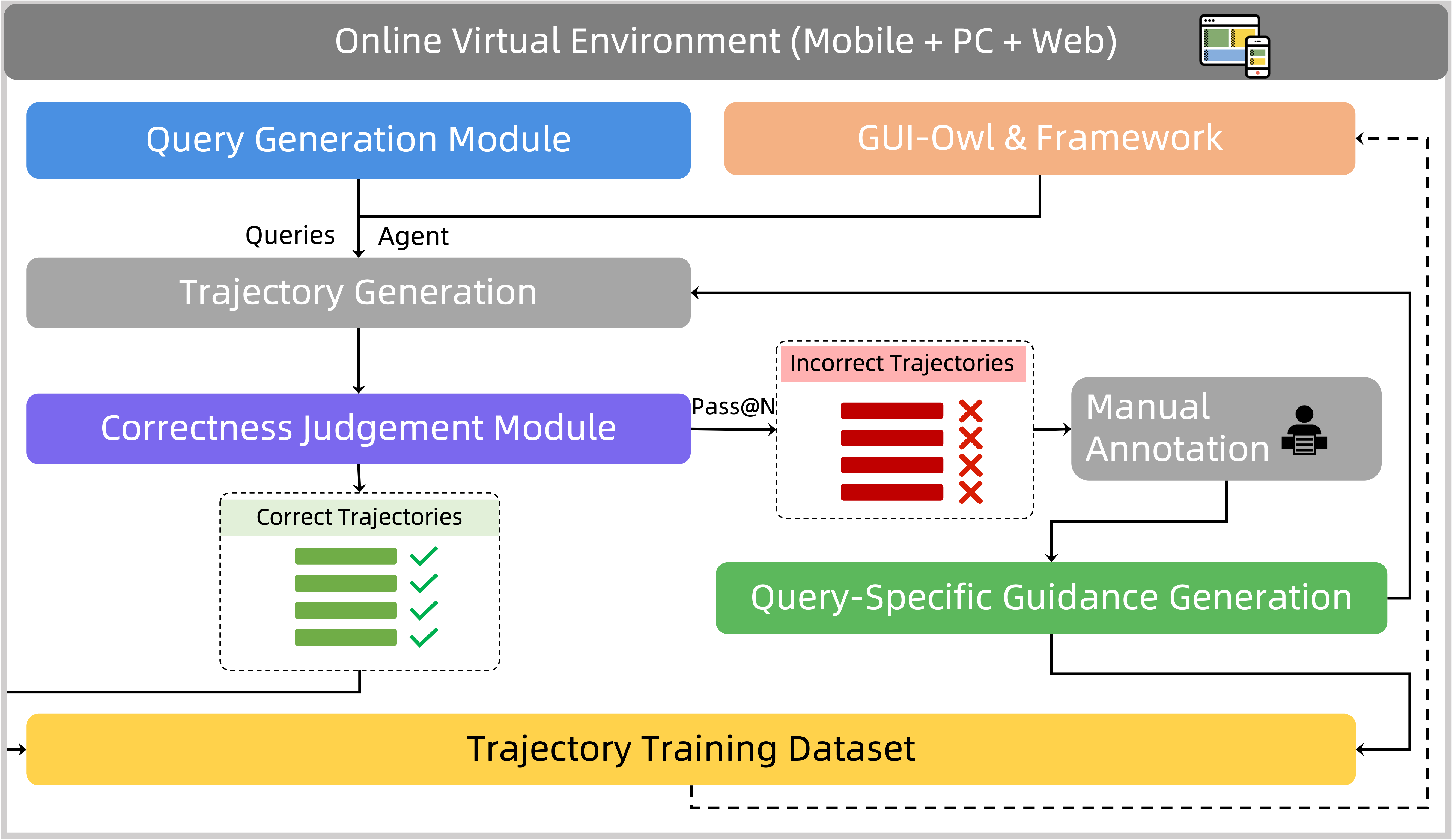}
    \caption{Illustration of the our self-evolving trajectory data production pipeline.}
    \label{fig:data_pipeline}
\end{figure*}

To scale up the trajectory data, we propose a Self-Evolving GUI Trajectory Production pipeline, which contrasts with traditional methods that strongly rely on manual annotation~\citep{wang2025opencuaopenfoundationscomputeruse,qin2025ui}.
This framework leverages
the capabilities of GUI-Owl itself, continuously generating new trajectories through roll-out and assessing their
correctness to obtain large-scale high-quality interaction data. Subsequently, these data are utilized to enhance the model’s
capabilities, creating a reinforcing cycle of improvement.
As shown in \Cref{fig:data_pipeline}, the process begins with constructing dynamic virtual environments across mobile, PC, and web platforms, paired with high-quality query generation. 
Given these queries, the GUI-Owl model and Mobile-Agent-v3 framework performs step-by-step actions in the environments to produce roll-out trajectories. A Trajectory Correctness Judgment Module then evaluates these trajectories at both the step and trajectory levels
to identify and filter out errors. For difficult queries, a Query-specific Guidance Generation module provides human- or model-generated ground-truth trajectories to guide the agent. The cleaned and enriched data is then used for reinforcement fine-tuning, enabling the model to iteratively improve its ability to generate successful GUI trajectories, thereby reducing human annotation needs and achieving continuous self-improvement. More details can be found in Section~\ref{section6}.

\begin{itemize}
 \item \textbf{High-Quality Query Generation}. 
 For mobile apps, we developed a screenshot-action framework utilizing a human-annotated Directed Acyclic Graph (DAG)~\citep{patil2025bfcl} that models realistic navigation flows and captures multi-constraint user queries. The process involves path sampling, metadata extraction, instruction synthesis using LLMs, refinement through few-shot LLM prompting, and interface validation via web crawlers. This framework minimizes LLM hallucinations while ensuring realistic and controllable query generation. For computer applications, we addressed the challenges of atomic operational skills and software operational pathways. We combined manual annotation and LLM-assisted generation to create queries for atomic operations (e.g., \textit{clicking, scrolling, dragging}) and complex software interactions. We utilized accessibility trees and deep-search chains to acquire operational pathways, and employed MLLMs to generate executable commands based on screenshots and exemplar inputs. This comprehensive approach ensures diverse, realistic, and accurate query generation across different GUI environments.
 
\item \textbf{Trajectory Correctness Judgment Module}. Our Trajectory Correctness Judgment Module employs a two-tiered system to evaluate the quality of generated GUI trajectories. It consists of a Step-Level Critic and a Trajectory-Level Critic. The Step-Level Critic analyzes individual actions within a trajectory by examining pre-action and post-action states, producing an analysis, summary, and categorical annotation (GOOD, NEUTRAL, HARMFUL) for each step. The Trajectory-Level Critic assesses the overall trajectory using a dual-channel approach: a Textual Reasoning Channel leveraging large language models, and a Multi-Modal Reasoning Channel incorporating both visual and textual data. The final correctness judgment is determined through a consensus mechanism.
This comprehensive approach ensures robust evaluation of GUI trajectories, combining granular step-level insights with holistic trajectory-level assessment to maintain high-quality training data for our GUI-Owl model.

\item \textbf{Query-specific Guidance Generation}. This module utilizes successful trajectories to create guidance for improved model performance. The process involves:
(1) Action Description: A VLM generates descriptions for each action's outcome in reference trajectories. Inputs include pre- and post-action screenshots and action decisions. For coordinate-based actions, we highlight interaction points to aid VLM analysis.
(2) Quality Control: For model-generated trajectories, the VLM cross-references the model's decision rationale to validate step effectiveness, filtering out suboptimal actions.
(3) Guidance Synthesis: Action descriptions are concatenated and fed into a Large Language Model (LLM), which summarizes the essential steps required to complete the query, producing query-specific guidance.
This approach enables the generation of targeted guidance, potentially improving the model's ability to handle complex queries and reducing the need for extensive rollouts or manual annotations.

\end{itemize}

\subsubsection{Diverse GUI Data Synthesis}
\begin{figure*}[htpb]
    \centering
    \includegraphics[width=0.99\textwidth]{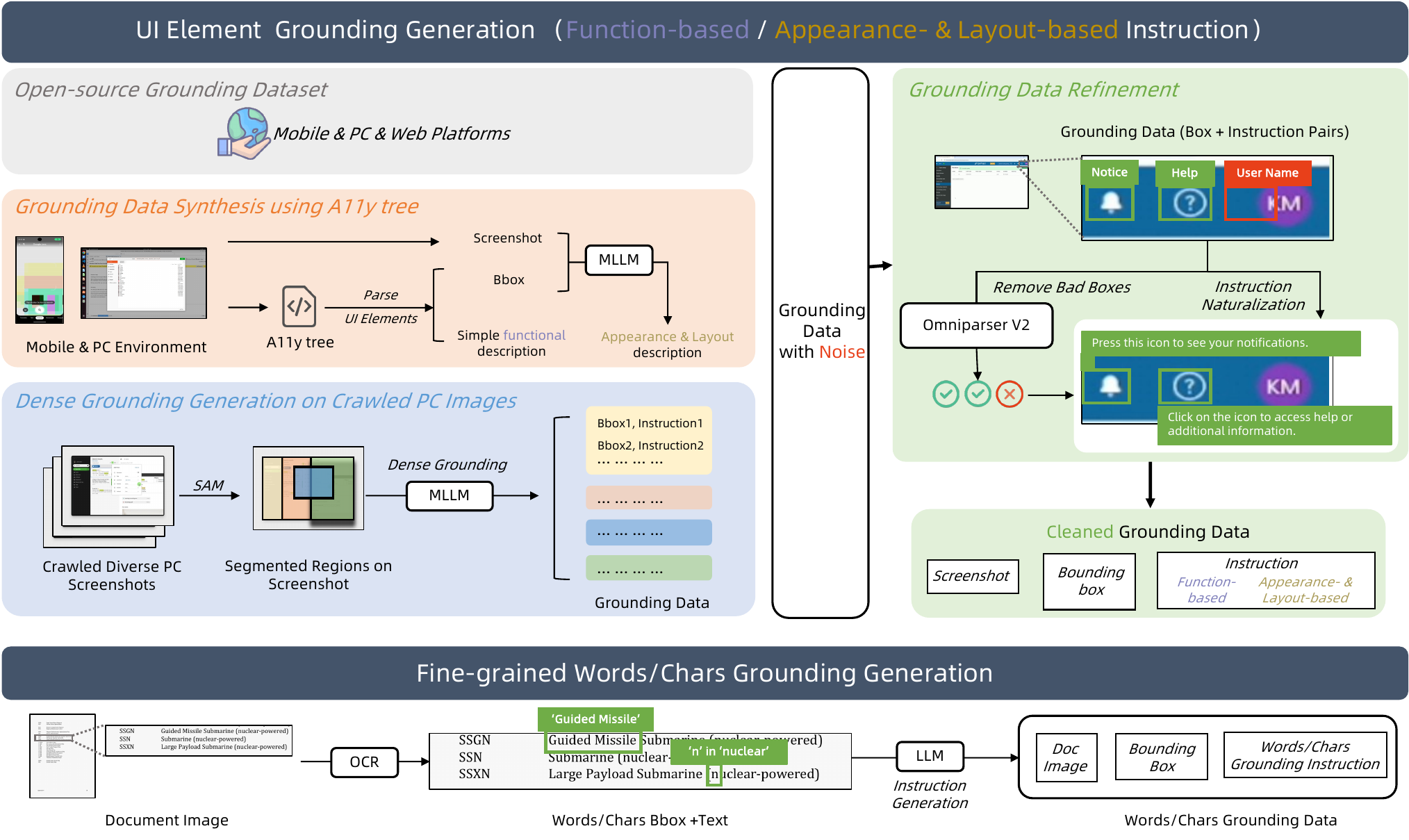}
    \caption{Overview of our grounding data construction pipeline.}
    \label{fig:ground_data}
\end{figure*}
\paragraph{Grounding.}
Accurate localization and semantic understanding of graphical user interface elements are essential for the development of robust and reliable visual interface agents, with these capabilities primarily embodied in grounding.
As illustrated in Figure~\ref{fig:ground_data}, to improve grounding capabilities, we construct two types of grounding task datasets from multiple data sources.

\begin{itemize}
\item For \textit{UI element grounding} (with \textit{function-based} or \textit{appearance- \& layout-based} instruction), we collect data from three sources: 
1) Open-source datasets:
Publicly released GUI datasets are utilized from UI-Vision~\citep{nayak2025ui}, and GUI-R1~\citep{luo2025gui}.
2) Grounding data synthesis using A11y tree:
 Extracting bounding boxes and functional descriptive information of UI elements through the accessibility (a11y) tree in both mobile and computer environments. And the appearance and layout descriptions are additionally generated using MLLMs such as Qwen2.5VL~\citep{Qwen2.5-VL}.
3) Dense grounding generation on crawled PC images:
To tackle the scarcity of PC grounding data, diverse screenshots are crawled from Google Images using popular app names as keywords. 
Given the high density and visual complexity of UI components in PC screenshots, we employ SAM~\citep{kirillov2023segany} to segment images into subregions, enabling more precise grounding. MLLMs then perform dense grounding within each segmented region.

To reduce noise and further improve the quality, we clean the collected grounding annotations by comparing them with UI detection results from Omniparser V2~\citep{yu2025omniparser} (bounding boxes with $IoU < \tau_g$ are removed, where $\tau_g = 0.5$).
Additionally, we rephrase the original instructions into more natural, task-oriented descriptions using LLMs (e.g., Qwen2.5-Max~\citep{qwen25}).

\item For \textit{fine-grained words and characters grounding}, we collect a set of document images and employ OCR tools to extract textual content and their corresponding spatial positions. Based on the annotated data, we build fine-grained text localization data to enable precise grounding of specific words and characters.
\end{itemize}

\paragraph{Task Planning.}
As foundational agents are often used to accomplish long-horizon, complex tasks, the model needs to possess background knowledge of complex task planning. We construct such data from two perspectives:

\begin{itemize}
\item 1) \textit{Distilling from Historical Trajectories.} Given historical successful trajectory data, we first construct fine-grained descriptions of each page transition. This information is then combined with the model’s historical actions and organized into a task execution manual through an LLM. By feeding this manual into GUI-Owl, we evaluate its quality based on changes in the task completion rate.

\item 2) \textit{Distilling from Large-scale Pretrained LLM.} To further enhance the model’s generalization on various tasks, we further distill knowledge from large-scale pretrained LLMs. First, we collect a list which covering mainstream apps, and then use either human annotators or models to synthesize plausible tasks. These tasks are designed to be as complex as possible and to span multiple features and even multiple applications; task specifications with obvious errors are filtered out. We then feed these tasks to a language model (e.g., Qwen3-235B) and further consolidate and clean the resulting plans, yielding task-specific planning data.
\end{itemize}

\paragraph{Action Semantics.}
We notice that a model’s ability to perceive how actions affect page-state changes is crucial. Based on collected trajectories, we extensively collect a large corpus of pre- and post-action screenshots and construct a two-tier dataset. At the first tier, we require the model to directly predict the intervening action—including its type and parameters—based on the before-and-after images; such data can be obtained directly from offline-collected trajectories. 

Subsequently, we ask the model to produce a natural-language description that covers both the executed action and its effects. To construct annotations for this data, we design a workflow that first generates an action description from the pre-action screenshot and the given action parameters (for coordinate-awared actions, the target location is drawn on the image to cue the model), using a multimodal model (e.g., Qwen-VL-Max). We then use the same multimodal model with the before-and-after images to analyze page changes and assess whether the changes are semantically consistent with the action. Through multiple rounds of voting, we retain the higher-scoring action descriptions.
\subsubsection{Enhanced Robust Reasoning}
Reasoning ability is essential for a fundamental agent, as it enables the model to move beyond merely imitating action sequences and instead capture the underlying logic that governs them. To this end, we first propose a set of diverse data synthesis strategies to enrich the variety of reasoning patterns. We then integrate reinforcement learning to further align these reasoning patterns with the dynamics of real-world environments.

\paragraph{Offline Hint-Guided Rejection Sampling.} We synthesize reasoning data via \emph{rejection sampling}. Specifically, given a collected trajectory
\[
T = \{(a_0, S_0), (a_1, S_1), \dots, (a_t, S_t)\},
\]
we prompt VLMs to generate reasoning content for each step based on its preceding history. The generated reasoning is then separated from the original context and used independently for action prediction. We evaluate the validity of each reasoning sequence by checking whether the predicted action matches the ground-truth action.

To encourage diversity in reasoning patterns, we adopt hints of different styles, for instance, requiring the model to follow a fixed chain-of-thought template or encouraging it to produce the simplest possible reasoning process. During this procedure, we observe that, for certain steps, the VLMs struggle to obtain actions consistent with the ground truth. For such cases, we first manually verify the correctness of the ground-truth action. If the action is deemed reasonable, we then feed it back to the VLMs to guide the generation of reasoning conclusions consistent with that action type.

\paragraph{Distillation from Multi-Agent Framework.} We note that even when style prompts are provided to encourage end-to-end reasoning generation, the model can still be influenced by certain inherent biases, which in turn limit reasoning diversity. In contrast, a multi-agent framework decomposes a single-step decision into the collaboration of multiple specialized roles, each approaching the current step from a different perspective. Since each agent focuses exclusively on its own subtask, it can more effectively avoid such biases. Motivated by this observation, we collect the outputs of individual agents from the Mobile-Agent-v3, and employ a large language model to integrate their diverse reasoning contents into a unified end-to-end reasoning output. The resulting reasoning content is paired with the action sequences obtained from Mobile-Agent-v3, forming the training dataset for reasoning.

\paragraph{Iterative Online Rejection Sampling.}
We observe that improving the base model's reasoning capability also enhances its ability to accomplish a wider range of tasks. Moreover, newly generated trajectory data can be further exploited for model training. Therefore, we adopt an \emph{iterative online rejection sampling} framework, in which our model rolls out trajectories on query data under two modes:

\begin{enumerate}
    \item \textbf{End-to-end generation}: the model directly generates reasoning and action predictions in an end-to-end fashion, which is used to improve its holistic reasoning capability.
    \item \textbf{Integration with Mobile-Agent-v3}: \modelname is incorporated into the Mobile-Agent-v3 framework. The inputs and outputs are collected to train the corresponding agent role models.
\end{enumerate}

Formally, given query data $\mathcal{Q}$ and a model $M^{(k)}$ at iteration $k$, trajectories are generated as:
\begin{equation}
    \mathcal{T}^{(k)} = \texttt{Rollout}\big(M^{(k)}, \mathcal{Q}\big),
\end{equation}
where $\mathcal{T}^{(k)}$ contains both end-to-end outputs $\mathcal{T}^{(k)}_{\text{E2E}}$ and role-specific outputs $\mathcal{T}^{(k)}_{\text{Role}}$.  
The newly collected trajectories are then used to update the model:
\begin{equation}
    M^{(k+1)} = \texttt{Train}\big(M^{(k)}, \mathcal{T}^{(k)}_{\text{filtered}}\big).
\end{equation}

Directly training on all collected steps often yields a suboptimal model. To address this, we apply the following filtering and balancing strategies:

\begin{enumerate}
    \item \textbf{Critic-based filtering}: A Critic pipeline scores each step $s_t \in \mathcal{T}^{(k)}$, and those with scores below a threshold $\tau_c$ are removed:
    \[
        \mathcal{T}_{\text{filtered}} = \{ s_t \mid \texttt{CriticScore}(s_t) \geq \tau_c \}.
    \]
    
    \item \textbf{Thought--action consistency check}: We verify the logical consistency between the reasoning content (thought) and the executed action. Steps that fail this check are discarded.
    
    \item \textbf{Task re-weighting}: Let $p_{\text{succ}}(task)$ denote the success rate of a given task.  
    For tasks with high $p_{\text{succ}}$, we downsample their training occurrence, while for tasks with low $p_{\text{succ}}$, we upsample their instances to ensure balanced learning.
    
    \item \textbf{Reflector balancing}: We observe that the Reflector Agent predominantly produces positive outputs, leading to class imbalance. We recalibrate its data as follows:
    \begin{itemize}
        \item If the Reflector marks step $i$ as negative and this feedback causes step $i{+}1$ to be judged positive, the feedback for step $i$ is retained.
        \item Otherwise, we retain Reflector inputs and responses only from trajectories where \emph{all} steps are judged positive by the reflactor.
    \end{itemize}
    Finally, we re-balance the dataset so that positive and negative samples have equal size.
\end{enumerate}

\section{Training Paradigm}
GUI-Owl is initialized from Qwen2.5-VL and trained through a three-stage process designed to progressively enhance its capabilities in GUI understanding, reasoning, and robust execution. 

\begin{itemize}
    \item \textbf{Pre-training Phase:} We collect a large-scale pre-training corpus covering fundamental UI understanding, interaction trajectory data, and general reasoning data. This data is used to continually pre-train Qwen2.5-VL, strengthening its grounding in basic GUI element recognition, action prediction, and general reasoning, thereby establishing a strong foundation for subsequent interaction-oriented training.
    
    \item \textbf{Iterative Tuning Phase:} We deploy the model in real-world environments such as desktops and mobile devices to perform large-scale task execution. The resulting trajectories are cleaned, scored, and further transformed into diverse reasoning datasets, which are then used for offline training. This iterative Tuning process enables GUI-Owl to accumulate effective reasoning patterns applicable across varied scenarios, improving its adaptability and decision-making in complex UI tasks.
    
    \item \textbf{Reinforcement Learning Phase:} We develop an asynchronous RL framework that allows the model to efficiently learn from direct interaction with real environments. This phase focuses on reinforcing successful behaviors and increasing execution consistency, thereby improving both the success rate and stability of GUI-Owl in practical deployments.
\end{itemize}

\subsection{Scalable Reinforcement Learning}

\begin{figure*}[!h]
    \centering
    \includegraphics[width=0.8\textwidth]{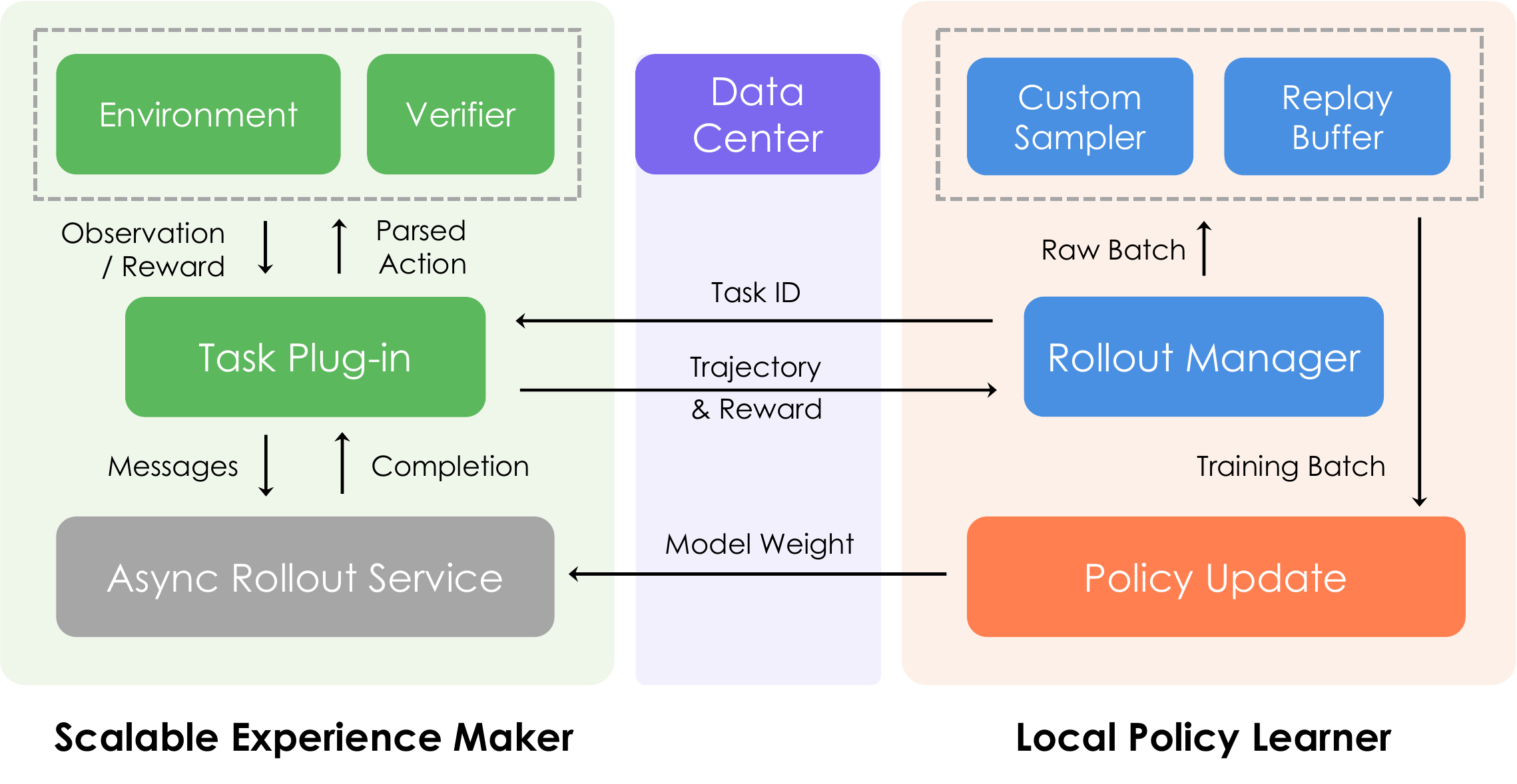}
    \caption{Overview of our scalable RL infrastructure, which unifies single-turn reasoning and multi-turn agentic training in a fully decoupled rollout–update framework. All components can run in parallel for high throughput, with diverse task-specific interactions plugged into the scalable experience maker with a unified interface. A rollout manager assigns task IDs, collects trajectories and rewards, and coordinates data flow via a shared data center.}
    \label{fig:rl_infra}
\end{figure*}

\subsubsection{Infrastructure}

Where enriched trajectory and reasoning data expand the model’s knowledge base and reasoning capabilities, the model is expected to exhibit lower uncertainty and higher stability in real-world usage. Therefore, we further introduce reinforcement learning to better align GUI-Owl with practical application.

To facilitate an efficient and flexible training framework for training with environment multi-turn interactions, we develop a general infrastructure with the following key features:




\noindent\makebox[1em][l]{\textbullet} \textbf{Unified Interface for multi-task training:} Our framework is built on a unified task plug-in interface that standardizes interactions for both single-turn reasoning and complex, multi-turn agentic tasks. This modular design allows diverse new environments and tasks to be seamlessly integrated without altering the core infrastructure.

\noindent\makebox[1em][l]{\textbullet} \textbf{Decoupled, Controllable Rollout:} We decouple the experience generation (rollout) phase from policy updates, giving operators precise control over the entire data supply chain. This control is multi-faceted: the manager can dictate the degree of policy-adherence, from a strictly synchronous on-policy mode to an asynchronous, slightly off-policy mode for speed. It also has full control over resource allocation, deploying rollouts on hardware optimized for inference to maximize throughput. This granular control enables us to fine-tune the data generation process to achieve an optimal balance among optimization guarantees, speed, and cost.

\subsubsection{Task Mixture}
We apply GRPO~\citep{guo2025deepseek} to train \modelname on static tasks, and apply the trajectory-aware Relative Policy Optimization (TRPO) strategy for training in online environments. In this section, we present the data preparation methods for different downstream reinforcement learning tasks and introduce trajectory-aware relative policy optimization.

For grounding task, a subset of data from GUI-R1~\citep{luo2025gui}, UI-Vision~\citep{nayak2025ui} serves as the foundational dataset. To further enhance RL performance in challenging fine-grained grounding, a curated collection of high-difficulty fine-grained grounding samples is incorporated (i.e., where the target UI regions occupy less than 0.1\% of the entire screenshot area).
Subsequently, the all datasets are performed $n_g$ sampling iterations (where $n_g=8$ in our implementation) utilizing the policy model GUI-Owl before RL, and sample instances that exhibit partial failure cases as training corpus for RL optimization.

To enhance the capabilities of low-level (i.e., step-level) actions, we introduce single-turn reinforcement learning. The training data is derived directly from individual steps within high-quality offline interaction trajectories.

While the preceding RL phases build foundational skills, applying them to complex, multi-step tasks in an online environment introduces significant challenges. Therefore,, we also conduct online reinforcement learning for GUI-Owl on virtual environments. These tasks are selected from the training task pool and use either rule-based or critic-based rewards as feedback signals for determining task completion.

\paragraph{Trajectory-aware Relative Policy Optimization for  Online Environment RL.}
\label{method:rl}

Real-world user tasks are often characterized by long and variable-length action sequences. In such scenarios, rewards are typically sparse and only available as a delayed success signal upon task completion. To overcome these obstacles, we employ a trajectory-aware relative policy optimization strategy (TRPO) extended to GRPO~\citep{guo2025deepseek}. This approach circumvents the challenge of assigning per-step rewards, a task that is nearly impossible to perform accurately in complex GUI interactions. Instead, we evaluate the entire trajectory $\tau$ after its completion. Our experiments are conducted on the OSWorld-Verified benchmark~\citep{xie2024osworld}, where the outcome of each task is programmatically verifiable, allowing us to obtain a reliable, single, holistic reward scalar $R(\tau)$. Specifically, this reward is the sum of an accuracy reward (1 for a successful trajectory, 0 otherwise) and a format reward, which penalizes malformed actions with a value of -0.5.

This trajectory-level reward is then used to compute a normalized advantage estimate, which provides a stable learning signal across all steps of the trajectory. The advantage for a given trajectory $\tau$ is calculated as: $\hat{A}_{\tau} = \frac{R(\tau) - \bar{R}}{\sigma_{R} + \epsilon},$
where $\bar{R}$ and $\sigma_{R}$ are the running mean and standard deviation of rewards observed across multiple trajectories. This normalized advantage $\hat{A}_{\tau}$ is then uniformly distributed to every action taken within that trajectory, ensuring that all steps contributing to a successful outcome receive a consistent positive signal, thereby mitigating the credit assignment problem.

Given the inherent sparsity of successful trajectories in GUI tasks, we incorporate a replay buffer to stabilize training. This buffer stores historically successful trajectories, indexed by their \texttt{task\_id}. During the sampling process, if a generated group of trajectories for a task results entirely in failures, one failing trajectory is randomly replaced with a successful one from the buffer corresponding to the same task. This injection of positive examples ensures the effectiveness of the training signal in every batch.


Our final policy optimization objective for a batch of $G$ trajectories is defined by the following loss function:
\begin{equation}
  \mathcal{L}_{\text{TRPO}} = -\frac{1}{N} \sum_{i=1}^{G} \sum_{s=1}^{S_i} \sum_{t=1}^{|\mathbf{o}_{i,s}|}
  \left\{
  \min \left[
  r_t(\theta)\hat{A}_{\tau_i}, \,
  \text{clip}\left(
  r_t(\theta),
  1 - \epsilon, 1 + \epsilon
  \right) \hat{A}_{\tau_i}
  \right]
  \right\}
\end{equation}
where $N$ is the total number of tokens in the batch, $\hat{A}_{\tau_i}$ is the trajectory-level advantage for trajectory $i$, and $r_t(\theta) = \frac{\pi_\theta(o_{s,t}|\dots)}{\pi_{\theta_{\text{old}}}(o_{s,t}|\dots)}$ is the probability ratio of a token under the current and old policies. This clipped objective function stabilizes training while effectively leveraging the holistic trajectory-level reward signal for long-horizon GUI automation tasks.

In practice, the high resolution of GUI screenshots means that a complete interaction trajectory can quickly exceed the model's context length (e.g., 32k for Qwen2.5-VL). To manage this, we segment each full multi-turn trajectory into several single-step data instances for the policy update. The loss computed for each step-wise instance is then scaled by the total number of steps in its original, complete trajectory. This approach addresses the issue of unbalanced optimization for trajectories of different lengths.
\section{Mobile-Agent-v3}
The agent-based framework method modularizes complex GUI tasks into multiple relatively simple tasks, and can achieve higher performance with the cooperation of agents with different roles.~\citep{zhang2025appagent,li2024appagent,wang2024mobile2,wang2025mobile,Agent-S2,agashe2024agent,zhang2025agentcpm,li2025mobileuse,liu2024autoglm,nong2024mobileflow,wu2024copilot,wu2024atlas,sun2024genesis,zhang2024ufo,zheng2024gpt,patel2024large,niu2024screenagent,tan2024cradle}. As noted earlier, GUI-Owl possesses multi-agent collaboration capabilities. Building upon this foundation, we further propose Mobile-Agent-v3, a multi-agent framework endowed with capabilities for knowledge evolution, task planning, sub-task execution, and reflective reasoning. In this section, we discuss the architecture of Mobile-Agent-v3.

\begin{figure*}[!ht]
    \centering
    \includegraphics[width=0.9\textwidth]{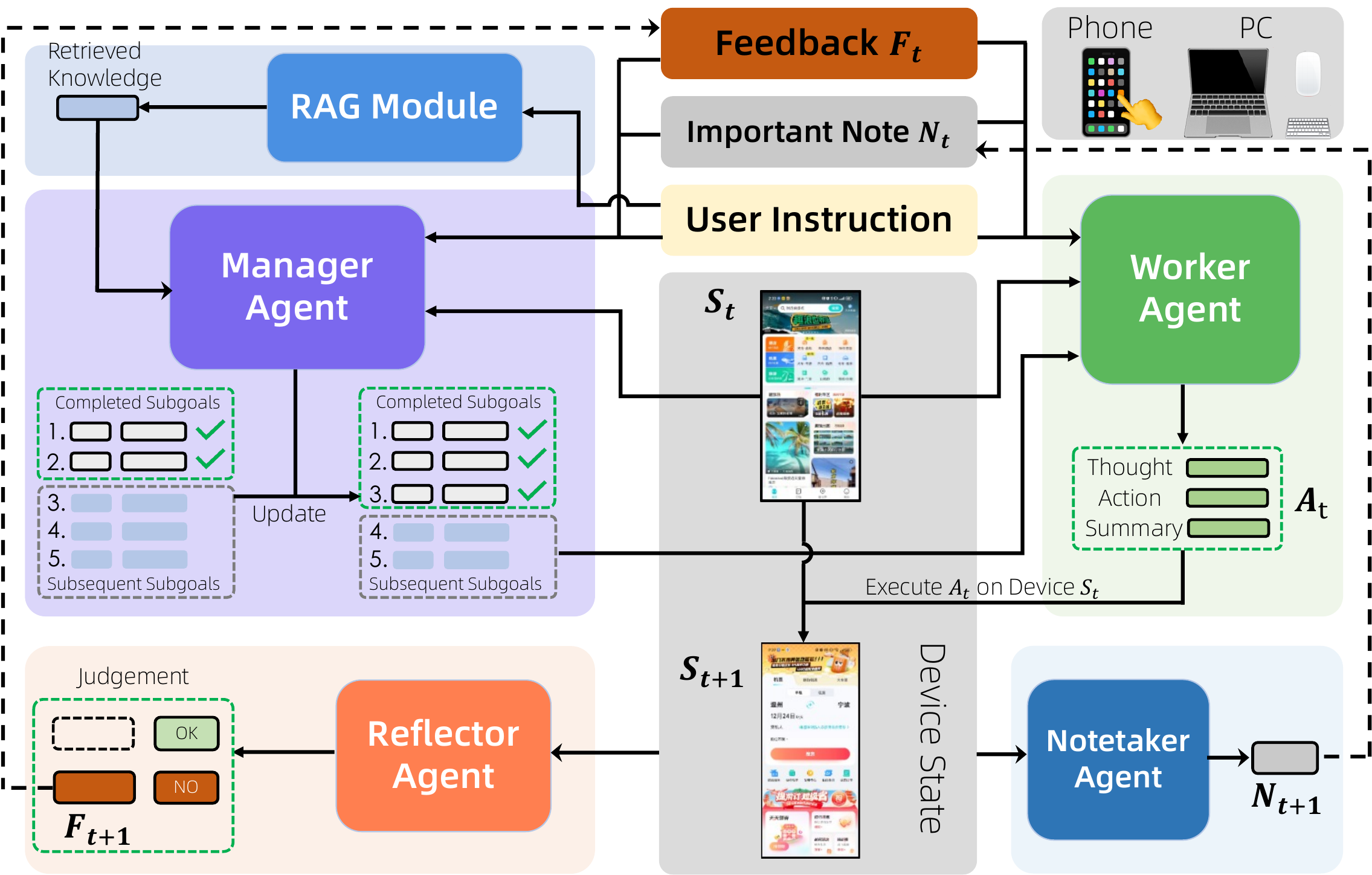}
    \caption{Mobile-Agent-v3 architecture. The system consists of six modules: (1) a RAG module for retrieving external world knowledge, (2) a Manager Agent for subgoal planning and guidance, (3) a Worker Agent for GUI operation, (4) a Reflector Agent for evaluation and feedback (5) a Notetaker Agent for recording important note, and (6) A GUI device interface supporting phone and PC environments.}
    \label{fig:gui_agent}
\end{figure*}

As presented in \Cref{fig:gui_agent}, the Mobile-Agent-v3 framework coordinates four specialized agents to achieve robust, adaptive, and long-horizon GUI task automation:

\begin{itemize}
    \item \textbf{Manager Agent} ($\mathcal{M}$): Serves as the strategic planner. At initialization, it decomposes a high-level instruction $I$ into an ordered subgoal list $SS_0$ using external knowledge $K_{\text{RAG}}$. During execution, it updates the plan based on results and feedback, re-prioritizing, modifying, or inserting corrective subgoals.
    
    \item \textbf{Worker Agent} ($\mathcal{W}$): Acts as the tactical executor. It selects and performs the most relevant actionable subgoal from $SS_t$ given the current GUI state $S_t$, prior feedback $F_{t-1}$, and accumulated notes $\mathcal{N}_t$, producing an action tuple $A_t$ that records reasoning, action, and intent.
    
    \item \textbf{Reflector Agent} ($\mathcal{R}$): Functions as the self-correction mechanism. It compares the intended outcome from the Worker with the actual state transition $(S_t \rightarrow S_{t+1})$, classifying the result as \texttt{SUCCESS} or \texttt{FAILURE} and generating detailed causal feedback $\phi_t$ for the Manager.
    
    \item \textbf{Notetaker Agent} ($\mathcal{C}$): Maintains persistent contextual memory. Triggered only on \texttt{SUCCESS}, it extracts and stores critical screen elements (e.g., codes, credentials) as notes $N_t$. The cumulative memory $\mathcal{N}_{t+1}$ supports both planning and execution in future steps.
\end{itemize}

The Manager decomposes and dynamically updates the plan, the Worker executes selected subgoals, the Reflector evaluates outcomes and provides diagnostic feedback, and the Notetaker preserves valuable transient information. This loop continues until all subgoals are completed or the instruction $I$ is fulfilled. More details can be found in Section~\ref{section7}.

\section{Experiments}
\subsection{Model Evaluation}

In this section, we evaluate \modelname on a wide range of benchmarks to thoroughly assess its performance as a fundamental agent in GUI-based scenarios. We train \modelname-7B and \modelname-32B, which are initialized from the Qwen2.5-VL models of the corresponding sizes. We conduct extensive experiments to evaluate \modelname\ in four key dimensions: grounding capability, comprehensive GUI understanding, end-to-end agent capability, and multi-agent capability. 

\subsubsection{grounding capability}

\begin{table*}[t]
\centering
\setlength{\tabcolsep}{10pt} 
\resizebox{0.9\textwidth}{!}{
\begin{tabular}{l rr rr rr >{\columncolor{lightgray}}r}
\toprule
 & \multicolumn{2}{c}{\textbf{Mobile}} & \multicolumn{2}{c}{\textbf{Desktop}} & \multicolumn{2}{c}{\textbf{Web}} &  \\
\cmidrule(lr){2-3} \cmidrule(lr){4-5} \cmidrule(lr){6-7}
\textbf{Agent Model} & \textbf{Text} & \textbf{Icon} & \textbf{Text} & \textbf{Icon} & \textbf{Text} & \textbf{Icon} & \textbf{Overall} \\
\midrule
\multicolumn{8}{l}{\textit{Proprietary Models}} \\ \midrule
Operator~\citep{cua2025}             & 47.3 & 41.5 & 90.2 & 80.3 & 92.8 & 84.3 & 70.5 \\
Claude 3.7 Sonnet~\citep{claude37}    & -    & -    & -    & -    & -    & -    & 87.6 \\
UI-TARS-1.5~\citep{qin2025ui}          & -    & -    & -    & -    & -    & -    & 94.2 \\
Seed-1.5-VL~\citep{seed2025seed1_5vl}         & -    & -    & -    & -    & -    & -    & 95.2 \\
\midrule
\multicolumn{8}{l}{\textit{Open-Source Models}} \\ \midrule
SeeClick~\citep{cheng2024seeclick}              & 78.4 & 50.7 & 70.1 & 29.3 & 55.2 & 32.5 & 55.1 \\
OmniParser-v2~\citep{yu2025omniparser}        & 95.5 & 74.6 & 92.3 & 60.9 & 88.0 & 59.6 & 80.7 \\
Qwen2.5-VL-3B~\citep{Qwen2.5-VL}       & 93.4 & 73.5 & 88.1 & 58.6 & 88.0 & 71.4 & 80.9 \\
UI-TARS-2B~\citep{qin2025ui}           & 95.2 & 79.1 & 90.7 & 68.6 & 87.2 & 78.3 & 84.7 \\
OS-Atlas-Base-4B~\citep{wu2024atlas}     & 95.2 & 75.8 & 90.7 & 63.6 & 90.6 & 77.3 & 85.1 \\
OS-Atlas-Base-7B~\citep{wu2024atlas}    & 96.2 & 83.4 & 89.7 & 69.3 & 94.0 & 79.8 & 87.1 \\
JEDI-3B~\citep{xie2025scalingcomputerusegroundinguser}              & 96.6 & 81.5 & 96.9 & 78.6 & 88.5 & 83.7 & 88.6 \\
Qwen2.5-VL-7B~\citep{Qwen2.5-VL}       & 97.6 & 87.2 & 90.2 & 74.2 & 93.2 & 81.3 & 88.8 \\
UI-TARS-72B~\citep{qin2025ui}         & 94.8 & 86.3 & 91.2 & 87.9 & 91.5 & 87.7 & 90.3 \\
UI-TARS-7B~\citep{qin2025ui}          & 96.9 & 89.1 & 95.4 & 85.0 & 93.6 & 85.2 & 91.6 \\
JEDI-7B~\citep{xie2025scalingcomputerusegroundinguser}              & 96.9 & 87.2 & 95.9 & 87.9 & 94.4 & 84.2 & 91.7 \\
\midrule
\modelname-7B       & 99.0 & 92.4 & 96.9 & 85.0 & 93.6 & 85.2 & \underline{92.8} \\
\modelname-32B       & 98.6 &90.0  & 97.9 &87.8  & 94.4 & 86.7 & \textbf{93.2} \\
\bottomrule
\end{tabular}
}
\caption{Comparison with state-of-the-art methods on the ScreenSpot-V2 dataset. \underline{Underlined} denotes the second-best open-source performance.}
\label{tab:ssv2}
\end{table*}


\begin{table*}[t]
\centering
\resizebox{\textwidth}{!}{%
\begin{tabular}{l rr rr rr rr rr rr >{\columncolor{lightgray}}r}
\toprule
& \multicolumn{2}{c}{\textbf{Development}} & \multicolumn{2}{c}{\textbf{Creative}} & \multicolumn{2}{c}{\textbf{CAD}} & \multicolumn{2}{c}{\textbf{Scientific}} & \multicolumn{2}{c}{\textbf{Office}} & \multicolumn{2}{c}{\textbf{OS}} & \textbf{Avg} \\
\cmidrule(lr){2-3} \cmidrule(lr){4-5} \cmidrule(lr){6-7} \cmidrule(lr){8-9} \cmidrule(lr){10-11} \cmidrule(lr){12-13} \cmidrule(lr){14-14}
\textbf{Agent Model} & \textbf{Text} & \textbf{Icon} & \textbf{Text} & \textbf{Icon} & \textbf{Text} & \textbf{Icon} & \textbf{Text} & \textbf{Icon} & \textbf{Text} & \textbf{Icon} & \textbf{Text} & \textbf{Icon} & \\
\midrule
\multicolumn{14}{l}{\textit{Proprietary Models}} \\ \midrule
GPT-4o~\citep{hurst2024gpt} & 1.3 & 0.0 & 1.0 & 0.0 & 2.0 & 0.0 & 2.1 & 0.0 & 1.1 & 0.0 & 0.0 & 0.0 & 0.8 \\
Claude 3.7 Sonnet~\citep{claude37}  & - & - & - & - & - & - & - & - & - & - & - & - & 27.7 \\
Operator~\citep{cua2025}   & 50.0 & 19.3 & 51.5 & 23.1 & 16.8 & 14.1 & 58.3 & 24.5 & 60.5 & 28.3 & 34.6 & 30.3 & 36.6 \\
Seed-1.5-VL~\citep{seed2025seed1_5vl} & - & - & - & - & - & - & - & - & - & - & - & - & 60.9 \\
UI-TARS-1.5~\citep{qin2025ui} & - & - & - & - & - & - & - & - & - & - & - & - & 61.6 \\
\midrule
\multicolumn{14}{l}{\textit{Open-Source Models}} \\ \midrule
UI-TARS-2B~\citep{qin2025ui}  & 47.4 & 4.1 & 42.9 & 6.3 & 17.8 & 4.7 & 56.9 & 17.3 & 50.3 & 17.0 & 21.5 & 5.6 & 27.7 \\
Qwen2.5-VL-3B~\citep{Qwen2.5-VL}  & 38.3 & 3.4 & 40.9 & 4.9 & 22.3 & 6.3 & 44.4 & 10.0 & 48.0 & 17.0 & 33.6 & 4.5 & 25.9 \\
Qwen2.5-VL-7B~\citep{Qwen2.5-VL}   & 51.9 & 4.8 & 36.9 & 8.4 & 17.8 & 1.6 & 48.6 & 8.2 & 53.7 & 18.9 & 34.6 & 7.9 & 27.6 \\
UI-R1-E-3B~\citep{lu2025ui}  &46.1& 6.9 &41.9 &4.2 &37.1 &12.5& 56.9& 21.8 &65.0 &26.4 &32.7& 10.1&33.5\\
UI-TARS-7B~\citep{qin2025ui}  & 58.4 & 12.4 & 50.0 & 9.1 & 20.8 & 9.4 & 63.9 & 31.8 & 63.3 & 20.8 & 30.8 & 16.9 & 35.7 \\
InfiGUI-R1-3B~\citep{liu2025infigui}  & 51.3 & 12.4 & 44.9 & 7.0 & 33.0 & 14.1 & 58.3 & 20.0 & 65.5 & 28.3 & 43.9 & 12.4 & 35.7 \\
JEDI-3B~\citep{xie2025scalingcomputerusegroundinguser}  & 61.0 & 13.8 & 53.5 & 8.4 & 27.4 & 9.4 & 54.2 & 18.2 & 64.4 & 32.1 & 38.3 & 9.0 & 36.1 \\
GUI-G1-3B~\citep{zhou2025gui}  & 50.7 & 10.3 & 36.6 & 11.9 & 39.6 & 9.4 & 61.8 & 30.0 & 67.2 & 32.1 & 23.5 & 10.6 & 37.1 \\
UI-TARS-72B~\citep{qin2025ui} & 63.0 & 17.3 & 57.1 & 15.4 & 18.8 & 12.5 & 64.6 & 20.9 & 63.3 & 26.4 & 42.1 & 15.7 & 38.1 \\
JEDI-7B~\citep{xie2025scalingcomputerusegroundinguser}  & 42.9 & 11.0 & 50.0 & 11.9 & 38.0 & 14.1 & 72.9 & 25.5 & 75.1 & 47.2 & 33.6 & 16.9 & 39.5 \\
Qwen2.5-VL-32B~\citep{Qwen2.5-VL}  & 74.0 & 21.4 & 61.1 & 13.3 & 38.1 & 15.6 & 78.5 & 29.1 & 76.3 & 37.7 & 55.1 & 27.0 & 47.6 \\
SE-GUI-7B~\citep{yuan2025enhancing} & 68.2 & 19.3 & 57.6 & 9.1 & 51.3 & 42.2 & 75.0 & 28.2 & 78.5 & 43.4 & 49.5 & 25.8 & 47.3 \\
GUI-G$^{2}$-7B~\citep{tang2025guig2} &68.8& 17.2& 57.1& 15.4 &55.8 &12.5& 77.1& 24.5& 74.0 &32.7& 57.9& 21.3& 47.5 \\
Qwen2.5-VL-72B~\citep{Qwen2.5-VL} & - & - & - & - & - & - & - & - & - & - & - & - & 53.3 \\
\midrule
\modelname-7B & 76.6 & 31.0 & 59.6 & 27.3 & 64.5 & 21.9 & 79.1 & 37.3 & 77.4 & 39.6 & 59.8 & 33.7 & \underline{54.9} \\
\modelname-32B & 84.4 & 39.3 & 65.2 & 18.2 & 62.4 & 28.1 & 82.6 & 39.1 & 81.4 & 39.6 & 70.1 & 36.0 & \textbf{58.0} \\
\bottomrule
\end{tabular}%
}
\caption{Comparison with state-of-the-art methods on the ScreenSpot-Pro dataset. \underline{Underlined} denotes the second-best open-source performance.}
\label{tab:ssp}
\end{table*}



\begin{table}[htbp]
\centering
\resizebox{0.95\textwidth}{!}{
\begin{tabular}{@{}lcccc>{\columncolor{lightgray}}c@{}}
\toprule
\textbf{Agent Model} & \textbf{Text Matching} & \textbf{Element Recog.} & \textbf{Layout Underst.} & \textbf{Fine-grained Manip.} & \textbf{Overall} \\
\midrule
\multicolumn{6}{l}{\textit{Proprietary Models}} \\ \midrule
Gemini-2.5-Pro~\citep{gemini25}    & 59.8 & 45.5 & 49.0 & 33.6 & 45.2 \\
Operator~\citep{cua2025}          & 51.3 & 42.4 & 46.6 & 31.5 & 40.6 \\
Seed1.5-VL~\citep{seed2025seed1_5vl}   & 73.9 & 66.7 & 69.6 & 47.0 & 62.9 \\\midrule
\multicolumn{6}{l}{\textit{Open-Source Models}} \\ \midrule
OS-Atlas-7B~\citep{wu2024atlas}      & 44.1 & 29.4 & 35.2 & 16.8 & 27.7 \\
UGround-V1-7B~\citep{gou2024navigating}    & 51.3 & 40.3 & 43.5 & 24.8 & 36.4 \\
Aguvis-7B~\citep{xu2024aguvis}        & 55.9 & 41.2 & 43.9 & 28.2 & 38.7 \\
UI-TARS-7B~\citep{qin2025ui}       & 60.2 & 51.8 & 54.9 & 35.6 & 47.5 \\
UI-TARS-72B~\citep{qin2025ui}       & 69.4 & 60.6 & 62.9 & 45.6 & \underline{57.1} \\
Qwen2.5-VL-3B~\citep{Qwen2.5-VL}    & 41.4 & 28.8 & 34.8 & 13.4 & 27.3 \\
Qwen2.5-VL-7B~\citep{Qwen2.5-VL}     & 45.6 & 32.7 & 41.9 & 18.1 & 31.4 \\
Qwen2.5-VL-32B~\citep{Qwen2.5-VL}    & 63.2 & 47.3 & 49.0 & 36.9 & 46.5 \\
JEDI-3B~\citep{xie2025scalingcomputerusegroundinguser}           & 67.4 & 53.0 & 53.8 & 44.3 & 50.9 \\
JEDI-7B~\citep{xie2025scalingcomputerusegroundinguser}           & 65.9 & 55.5 & 57.7 & 46.9 & 54.1 \\ 
\midrule
\modelname-7B & 64.8 & 63.6 & 61.3 & 41.0 & 55.9 \\
\modelname-32B & 67.0 & 64.5 & 67.2 & 45.6 & \textbf{58.0}\\
\bottomrule
\end{tabular}
}
\caption{Comparison with state-of-the-art methods on the OSWorld-G dataset. \underline{Underlined} denotes the second-best open-source performance.
}
\label{tab:oswg}
\end{table}




\begin{table}[ht]
\centering

\resizebox{\textwidth}{!}{%
\begin{tabular}{l rrrrrrrrrrrr>{\columncolor{lightgray}}r}
\toprule
\multirow{2}{*}{\textbf{Model}} & \multicolumn{2}{c}{\textbf{Windows}} & \multicolumn{2}{c}{\textbf{MacOS}} & \multicolumn{2}{c}{\textbf{Linux}} & \multicolumn{2}{c}{\textbf{iOS}} & \multicolumn{2}{c}{\textbf{Android}} & \multicolumn{2}{c}{\textbf{Web}} & \multirow{2}{*}{\textbf{Overall}} \\
\cmidrule(lr){2-3} \cmidrule(lr){4-5} \cmidrule(lr){6-7} \cmidrule(lr){8-9} \cmidrule(lr){10-11} \cmidrule(lr){12-13}
& Basic & Adv. & Basic & Adv. & Basic & Adv. & Basic & Adv. & Basic & Adv. & Basic & Adv. & \\
\midrule
GPT-4o~\citep{hurst2024gpt} & 1.48 & 1.10 & 8.69 & 4.34 & 1.05 & 1.02 & 5.10 & 3.33 & 2.53 & 1.41 & 3.23 & 2.92 & 2.87 \\
Claude-3.7~\citep{claude37}  & 1.48 & 0.74 & 12.46 & 7.51 & 1.05 & 0.00 & 13.69 & 10.61 & 1.40 & 1.40 & 3.23 & 2.27 & 4.66 \\
Qwen-Max-VL~\citep{Qwen2.5-VL} & 43.91 & 36.76 & 58.84 & 56.07 & 53.93 & 30.10 & 77.39 & 59.09 & 79.49 & 70.14 & 74.84 & 58.77 & 58.03 \\
Aguvis-7B-720P~\citep{xu2024aguvis}  & 37.27 & 21.69 & 48.12 & 33.27 & 33.51 & 25.00 & 67.52 & 65.15 & 60.96 & 50.99 & 61.61 & 45.45 & 45.66 \\
ShowUI-2B~\citep{lin2025showui}  & 9.23 & 4.41 & 24.06 & 10.40 & 25.13 & 11.73 & 28.98 & 19.70 & 17.42 & 8.73 & 22.90 & 12.66 & 15.96 \\
OS-Atlas-Base-7B~\citep{wu2024atlas}  & 36.90 & 18.75 & 44.35 & 21.68 & 31.41 & 13.27 & 74.84 & 48.79 & 69.60 & 46.76 & 61.29 & 35.39 & 41.42 \\
UGround-V1-7B~\citep{gou2024navigating} & 66.79 & 38.97 & 71.30 & 48.55 & 56.54 & 31.12 & 92.68 & 70.91 & 93.54 & 70.99 & 88.71 & 64.61 & 65.68 \\
InternVL3-72B~\citep{zhu2025internvl3} & 70.11 & 42.64 & 75.65 & 52.31 & 59.16 & 41.33 & 93.63 & 80.61 & 92.70 & 78.59 & 90.65 & 65.91 & 72.20 \\
Qwen2.5-VL-72B~\citep{Qwen2.5-VL} & 55.72 & 33.82 & 49.86 & 30.06 & 40.31 & 20.92 & 56.05 & 28.18 & 55.62 & 25.35 & 68.39 & 45.78 & 41.83 \\
Qwen2.5-VL-7B~\citep{Qwen2.5-VL}  & 31.37 & 16.54 & 31.30 & 21.97 & 21.47 & 12.24 & 66.56 & 55.15 & 35.11 & 35.21 & 40.32 & 32.47 & 33.85 \\
UI-TARS-1.5-7B~\citep{qin2025ui}  & 68.27 & 38.97 & 68.99 & 44.51 & 64.40 & 37.76 & 88.54 & 69.39 & 90.45 & 69.29 & 80.97 & 56.49 & 64.32 \\
UI-TARS-72B-DPO~\citep{qin2025ui}  & 78.60 & 51.84 & 80.29 & 62.72 & 68.59 & 51.53 & 90.76 & 81.21 & 92.98 & 80.00 & 88.06 & 68.51 & 74.25 \\ \midrule
\modelname-7B & 86.35 & 61.76 & 81.74 & 64.45 & 74.35 & 61.73 & 94.90 & 83.03 &  95.78 & 83.66 & 93.22 & 72.72 & \underline{80.49} \\
\modelname-32B & 85.61 & 65.07 & 84.93 & 67.05 & 76.96 & 63.27 & 95.22 & 85.45 & 96.07 & 87.04 & 95.48 & 80.84 & \textbf{82.97} \\
\bottomrule
\end{tabular}%
}
\caption{Comparison with state-of-the-art methods on the MMBench-GUI-L2 dataset. \underline{Underlined} denotes the second-best open-source performance.}
\label{tab:mmbench_l2}
\end{table}



The grounding capability evaluates a model’s ability to locate the corresponding UI element given a natural-language query. We use ScreenSpot V2, ScreenSpot Pro, OSWorld-G, and MMBench-GUI L2 as benchmarks. ScreenSpot v2 covers mobile, desktop, and web scenarios, while ScreenSpot-Pro primarily evaluates a model’s localization ability at ultra-high resolutions. OSWorld-G contains finely annotated queries. MMBench-GUI L2 has the broadest coverage and more faithfully reflects a model’s grounding performance in real-world settings. The performance comparisons are shown in \Cref{tab:ssv2}, \Cref{tab:ssp}, \Cref{tab:oswg} and \Cref{tab:mmbench_l2}. 

\modelname-7B achieves state-of-the-art performance among all 7B models. On screenspot-pro, which focuses on high-resolution images, we achieve a score of 54.9, significantly exceeding the performance of UI-TARS-72B and Qwen2.5-VL 72B. \modelname-7B also achieves competitive performance on OSWorld-G compared to UI-TARS-72B. \modelname-32B surpasses all models of the same size. MMBench-GUI-L2 evaluates a very broad and challenging set of queries, where our model scores 80.49, substantially outperforming all existing models. \modelname-32B further achieves a performance level of 82.97 and demonstrates leading grounding capabilities across various domains.

\subsubsection{Comprehensive GUI Understanding}

\begin{table*}[t]
\centering
\resizebox{0.8\textwidth}{!}{%
\begin{tabular}{lccccccc}
\toprule
\textbf{Model} & \textbf{Windows} & \textbf{MacOS} & \textbf{Linux} & \textbf{iOS} & \textbf{Android} & \textbf{Web} & \textbf{Overall} \\
\midrule
\multicolumn{8}{c}{\cellcolor{gray!20}\textit{Easy Level}} \\
GPT-4o~\citep{hurst2024gpt}  & 62.47 & 67.89 & 62.38 & 58.52 & 56.41 & 58.51 & 60.16 \\
Claude-3.5~\citep{claude35}  & 41.34 & 50.04 & 41.61 & 42.03 & 38.96 & 41.79 & 41.54 \\
Qwen2.5-VL-72B~\citep{Qwen2.5-VL}  & 65.86 & 75.23 & 73.02 & 67.24 & 58.09 & 72.08 & 66.98 \\
UI-TARS-72B-DPO~\citep{qin2025ui}  & 41.59 & 28.52 & 35.16 & 31.08 & 52.25 & 35.33 & 40.18 \\
InternVL3-72B~\citep{zhu2025internvl3} & 74.67 & 78.72 & 79.16 & 83.57 & 80.10 & 81.18 & 79.15 \\ \midrule
GUI-Owl-7B & 82.96 & 84.52 & 85.57 & 82.61 & 83.28 & 88.13 & 84.50 \\
GUI-Owl-32B & 93.70 & 89.29 & 93.30 & 95.65 & 90.49 & 94.06 & 92.75 \\

\midrule
\multicolumn{8}{c}{\cellcolor{gray!20}\textit{Medium Level}} \\
GPT-4o~\citep{hurst2024gpt}  & 56.33 & 63.13 & 59.70 & 54.06 & 57.69 & 54.98 & 57.24 \\
Claude-3.5~\citep{claude37}  & 39.28 & 47.63 & 45.97 & 44.57 & 42.03 & 34.33 & 41.26 \\
Qwen2.5-VL-72B~\citep{Qwen2.5-VL}  & 66.29 & 72.73 & 72.63 & 59.27 & 66.24 & 68.24 & 67.45 \\
UI-TARS-72B-DPO~\citep{qin2025ui}  & 38.83 & 41.60 & 37.14 & 41.72 & 54.74 & 31.55 & 41.77 \\
InternVL3-72B~\citep{zhu2025internvl3}  & 71.46 & 78.58 & 79.88 & 78.43 & 81.36 & 78.67 & 77.89 \\\midrule
GUI-Owl-7B & 88.89 & 88.10 & 91.24 & 84.35 & 85.25 & 83.56 & 86.86 \\
GUI-Owl-32B & 94.07 & 84.52 & 95.88 & 87.83 & 92.79 & 88.58 & 91.74 \\

\midrule
\multicolumn{8}{c}{\cellcolor{gray!20}\textit{Hard Level}} \\
GPT-4o~\citep{hurst2024gpt}  & 60.69 & 60.38 & 52.42 & 45.27 & 50.93 & 50.83 & 53.49 \\
Claude-3.5~\citep{claude37}  & 37.40 & 42.70 & 34.07 & 40.86 & 36.96 & 38.11 & 37.55 \\
Qwen2.5-VL-72B~\citep{Qwen2.5-VL}  & 70.68 & 68.91 & 70.98 & 57.59 & 53.94 & 68.10 & 64.56 \\
UI-TARS-72B-DPO~\citep{qin2025ui}  & 31.48 & 35.87 & 24.19 & 36.33 & 58.13 & 19.94 & 35.78 \\
InternVL3-72B~\citep{zhu2025internvl3}& 75.08 & 77.44 & 76.19 & 70.37 & 75.73 & 78.11 & 75.70 \\\midrule
GUI-Owl-7B & 87.78 & 96.43 & 94.33 & 87.83 & 88.85 & 94.06 & \underline{90.90} \\
GUI-Owl-32B & 93.33 & 95.24 & 95.88 & 92.17 & 95.41 & 92.69 & \textbf{94.19} \\

\bottomrule
\end{tabular}%
}
\caption{Comparison with state-of-the-art methods on the MMBench-GUI-L1 dataset. \underline{Underlined} denotes the second-best open-source performance.}
\label{tab:mmbench_l1}
\end{table*}

Comprehensive GUI Understanding examines whether a GUI model can accurately interpret interface states and produce appropriate responses. We adopt two benchmarks for this evaluation. MMBench-GUI-L1 assesses the model’s UI understanding and single-step decision-making capability through a question-answering format. Android Control evaluates the model’s ability to perform single-step decisions within pre-annotated trajectory contexts.

On the MMBench-GUI-L1 benchmark, GUI-Owl scores 84.5, 86.9, and 90.9 on the easy, medium, and hard levels, respectively, substantially outperforming all existing models. On Android Control, it achieves a score of 72.8, establishing the highest performance among all 7B models. We find that GUI-Owl-32B achieves a score of 76.6, surpassing the current state-of-the-art UI-TARS-72B. GUI-Owl-32B significantly outperforms GUI-Owl-7B across different difficulty levels and domains, reflecting its more comprehensive and sufficient reserve of GUI knowledge.

\subsubsection{End2end and Multi-Agent capability on Online environment}


\begin{table*}[htbp] 
\centering
\begin{minipage}{0.4\linewidth}
    \centering
    \resizebox{1\textwidth}{!}{%
    \begin{tabular}{lc}
    \toprule
    \textbf{Model} & \textbf{Score} \\
    \midrule
    Claude-3.5~\citep{claude35}            & 12.5  \\
    GPT-4o~\citep{hurst2024gpt}            & 20.8  \\
    Gemini 2.0~\citep{gemini2}        & 28.5  \\
    Qwen2-VL-72B~\citep{Qwen2-VL}      & 59.1  \\
    Aguvis-72B~\citep{xu2024aguvis}        & 66.4  \\
    Qwen2.5-VL-72B~\citep{Qwen2.5-VL}    & 67.4 \\
    UI-TARS-7B~\citep{qin2025ui}         & 72.5  \\
    UI-TARS-72B~\citep{qin2025ui}        & \underline{74.7}  \\ \midrule
    GUI-Owl-7B        & 72.8  \\
   GUI-Owl-32B        & \textbf{76.6} \\
    \bottomrule
    \end{tabular}
    }
    \caption{Model performance on the Android Control benchmark. 
    Extract match scores with high-level instruction are reported. \underline{Underlined} denotes the second-best open-source performance.}
    \label{tab:android_control}
\end{minipage}
\hfill
\begin{minipage}{0.56\linewidth}
    \centering
    \resizebox{1\textwidth}{!}{%
    \begin{tabular}{l rr rr}
    \toprule
     & \multicolumn{2}{c}{\textbf{Online}} \\
     \cmidrule(lr){2-3}
    \textbf{Agent Model} & \textbf{OSWorld-Verified} & \textbf{AndroidWorld} \\
    \midrule
    \multicolumn{3}{l}{\textit{Proprietary Models}} \\ \midrule
     SeedVL-1.5~\citep{seed2025seed1_5vl} & 34.1 & 62.1 \\
     Claude-4-sonnet~\citep{claude4} & 43.9  & - \\
     OpenAI CUA o3~\citep{openAI_o3_o4_mini} & 23.0 & - \\
     UI-TARS-1.5~\citep{qin2025ui} & - & 64.2 \\
    \midrule
    \multicolumn{3}{l}{\textit{Open-Source Models}} \\ \midrule
     UI-TARS-72B-DPO~\citep{qin2025ui} & 24.0 & 46.6 \\

     OpenCUA-7B~\citep{wang2025opencuaopenfoundationscomputeruse} & 28.2 & - \\
    OpenCUA-32B~\citep{wang2025opencuaopenfoundationscomputeruse} & 34.8 & - \\
     UI-TARS1.5-7B~\citep{qin2025ui} & 27.4 & - \\
    \midrule 
    \modelname-7B &  \underline{34.9}\textsuperscript{*} & \underline{66.4} \\
    Mobile-Agent-v3 & \textbf{37.7} & \textbf{73.3} \\
    \bottomrule
    \end{tabular}
    }
    \caption{Online evaluation results on OSWorld-Verified and AndroidWorld benchmarks. \underline{Underlined} denotes the second-best open-source performance.}
    \footnotetext{*A variant of \modelname specifically RL-tuned for a desktop environment (\Cref{sec:online_rl}). The general version of \modelname achieves a score of 29.4.}
    
    \label{tab:online_eval}
\end{minipage}
\end{table*}

While the aforementioned evaluations measure a model’s performance in single-step decision-making, they suffer from two main limitations: (1) Errors in individual steps do not accumulate, making it impossible to assess the ability to accomplish complete tasks; (2) Although there may be multiple valid ways to complete a task, the ground-truth step sequences may reflect specific preferences, which can result in an unfair evaluation across models. To more comprehensively evaluate both the end-to-end agent capability and the multi-agent capability, we adopt realistic interactive environments — AndroidWorld and OSWorld-Verified.

GUI-Owl-7B outperforms UITARS 1.5 on AndroidWorld, and Mobile-Agent-v3 achieves an even greater lead, significantly surpassing all existing models. On OSWorld-Verified, GUI-Owl also outperforms the open-source OpenCUA-7B. We further adopt \modelname-32B into Mobile-Agent-v3, it achieves 37.7 on OSWorld-Verified and 73.3 on AndroidWorld. This suggests that \modelname is not only capable of independently solving tasks, but is also well-suited for integration into a multi-agent framework.

\subsection{Trajectory-level Online Reinforcement Learning}
\label{sec:online_rl}
\begin{figure}[!h]
    \centering
    \includegraphics[width=0.7\textwidth]{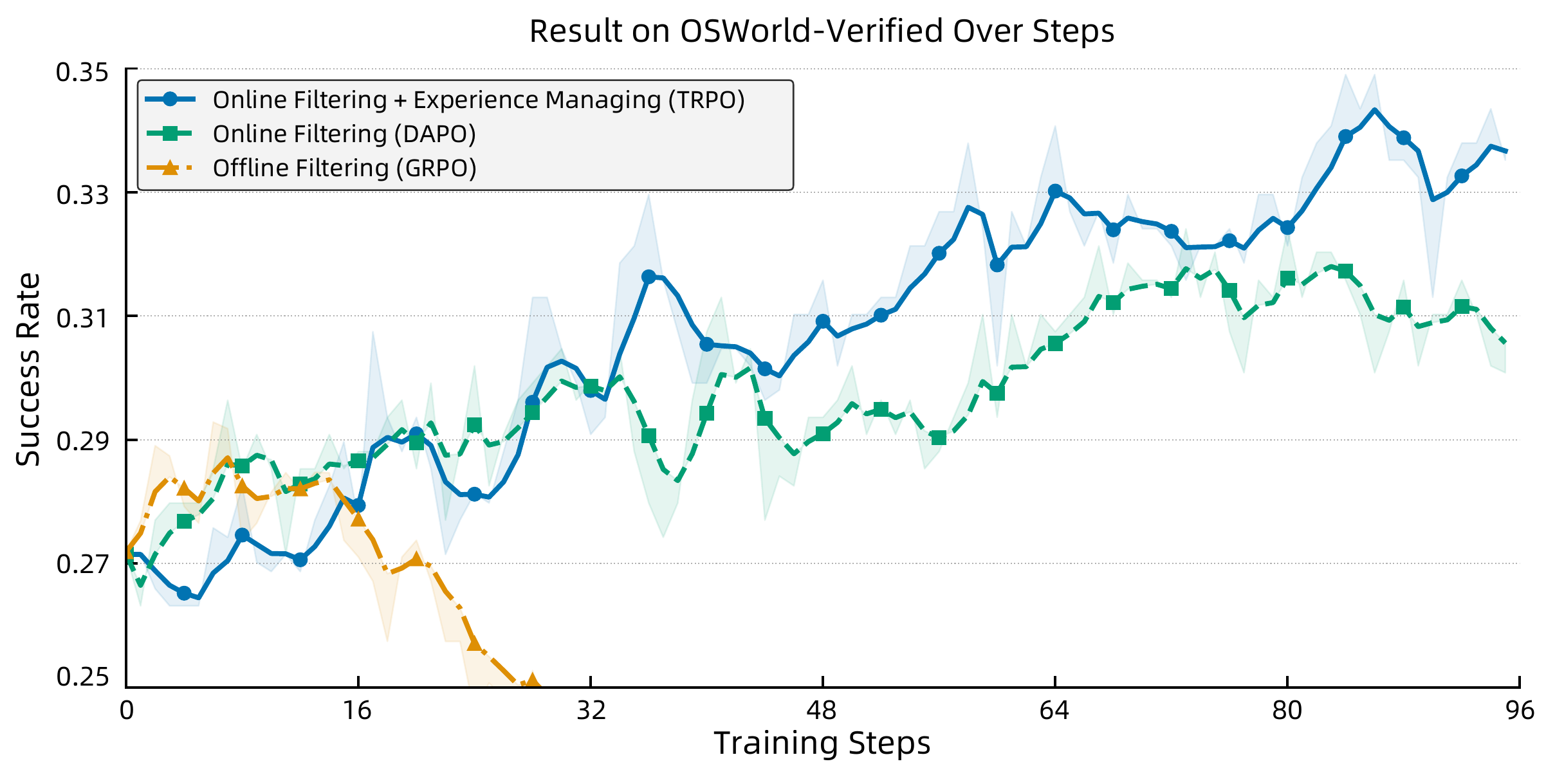}
    \caption{Training dynamics of GUI-Owl-7B on OSWorld-Verified. We limit the maximum interaction steps to 15 by default. \textit{Offline Filtering} removes tasks with all-success or all-failure outcomes before applying vanilla GRPO, serving as common preprocessing. \textit{Online Filtering} moves all tasks to online training and applies DAPO for selective filtering. \textit{Experience Managing} activates both the replay buffer and the use of leftover rollouts after batch filling, as described in Section~\ref{method:rl}.}
    \label{fig:rl_exp}
\end{figure}

To validate the efficacy of our trajectory-level online reinforcement learning strategy, we conducted a series of experiments on the OSWorld-Verified~\citep{xie2024osworld} benchmark, with all tasks limited to a maximum of 15 steps. The results, illustrated in Figure \ref{fig:rl_exp}, demonstrate the clear advantages of our proposed approach. Starting from an initial checkpoint with a 27.1 success rate, our method shows consistent, stable improvement throughout training, ultimately achieving a peak success rate of over 34.9. This steady learning curve underscores the effectiveness of our trajectory-aware relative policy optimization. By calculating a single, normalized advantage estimate $\hat{A}_{\tau}$ for an entire trajectory, our method successfully mitigates the severe credit assignment problem inherent in long-horizon GUI tasks and provides a coherent learning signal.

The most critical insight comes from the ablation study, \textit{Online Filtering (DAPO)}. This variant, which disables our successful-trajectory replay buffer and the mechanism for carrying over unused rollouts, confirms the value of our specific design choices. While this model still shows a positive learning trend, its performance is notably more volatile and ultimately inferior, peaking at around 31.5\% before declining. This instability highlights the challenge of sparse positive feedback; without the replay buffer injecting successful examples, the agent struggles to learn from the vast space of failing trajectories. The final performance gap between our full model and this ablation underscores the importance of data efficiency. By retaining and reusing all generated rollouts, our full method maximizes the utility of costly interactions, providing a richer training signal that leads to more stable and superior final performance.

Our comparison with the \textit{Offline Filtering (GRPO)} baseline further justifies our online data selection methodology. Offline filtering is a very common technique for preparing RL data by removing tasks that are statically identified as all-successful or all-failing across multiple inference runs. However, the results show this approach is not suitable for GUI automation tasks that require long-range, multi-step planning. After an initial small gain, its performance stagnates around a 29.1 success rate before degrading significantly. The failure arises because the final reward depends on a long sequence of actions, making outcomes highly sensitive to minor policy changes during training. Such sensitivity causes abrupt, non-linear shifts in success rates, in contrast to the smoother improvement observed in single-step reasoning tasks. Relying solely on offline filtering further aggravates this issue, leading to severe overfitting. As our results confirm, a more effective solution is a dynamic online filtering strategy, which continuously adapts the training distribution to the agent’s evolving policy.

In summary, the results validate that while trajectory-level optimization provides a solid foundation, it is our novel experience management, which combines a success-replay mechanism with maximum data utilization, that is crucial for achieving stable and efficient performance. This methodology allows GUI-Owl-7B to achieve state-of-the-art results among open-source models of the same model size. Notably, under identical experimental settings, our model also surpasses the performance of powerful proprietary models like Claude-4-Sonnet. This demonstrates that our specialized online RL fine-tuning strategy can effectively elevate strong base models, enabling them to excel in complex, long-horizon interactive tasks and rival the capabilities of significantly larger systems.

\subsubsection{Scaling of interaction steps and historical images}

\begin{figure}[!h]
    \centering
    \includegraphics[width=0.7\textwidth]{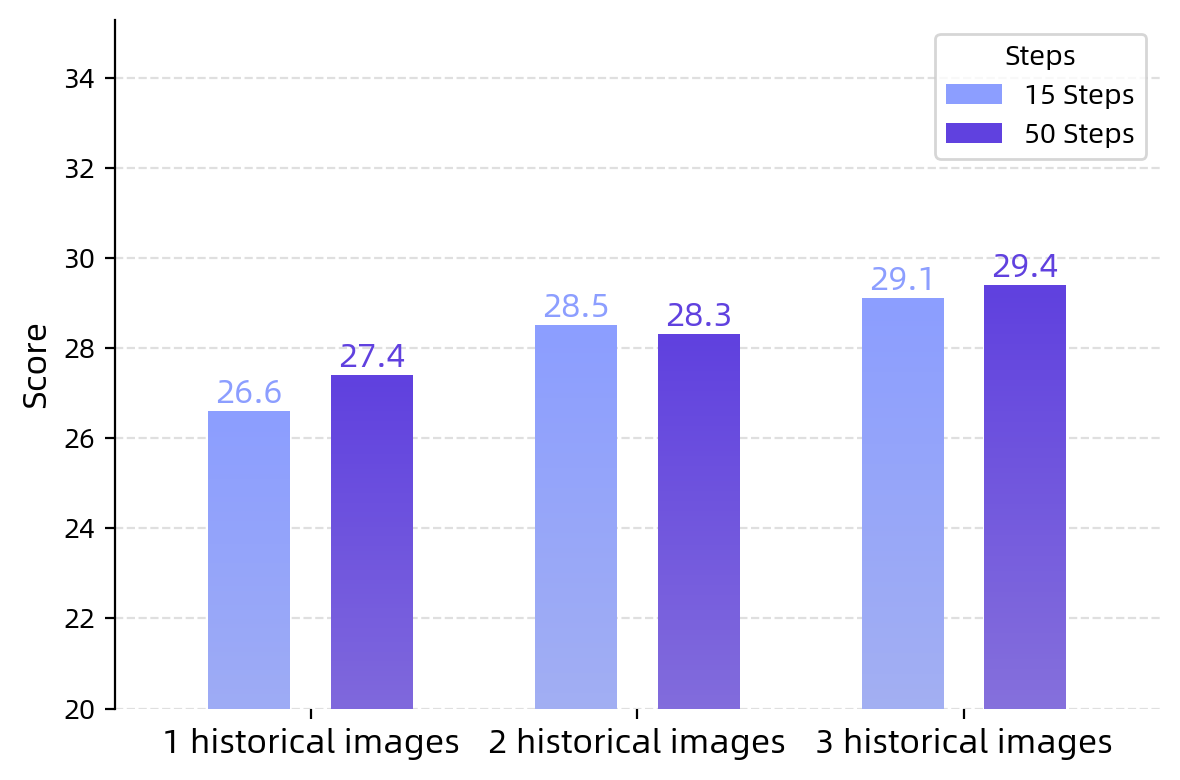}
    \caption{Performance of \modelname-7B on OSWorld-Verified with varying numbers of historical images and interaction-step budgets.}
    \label{fig:scaling_image_steps}
\end{figure}

We further analyze, on OSWorld, how GUI-Owl’s performance varies with the number of historical screenshots and the interaction-step budget. As shown in \Cref{fig:scaling_image_steps}, performance increases steadily as more historical images are provided. This is because the model’s understanding of UI changes relies on contrasts between consecutive frames, and additional images also help the model promptly reflect on and correct persistent erroneous behaviors. We also observe that increasing the interaction-step budget improves performance, indicating that our model has a significant advantage on long-horizon tasks.

\subsection{Effect of Reasoning Data Synthesis}
\begin{figure}[!h]
    \centering
    \includegraphics[width=\textwidth]{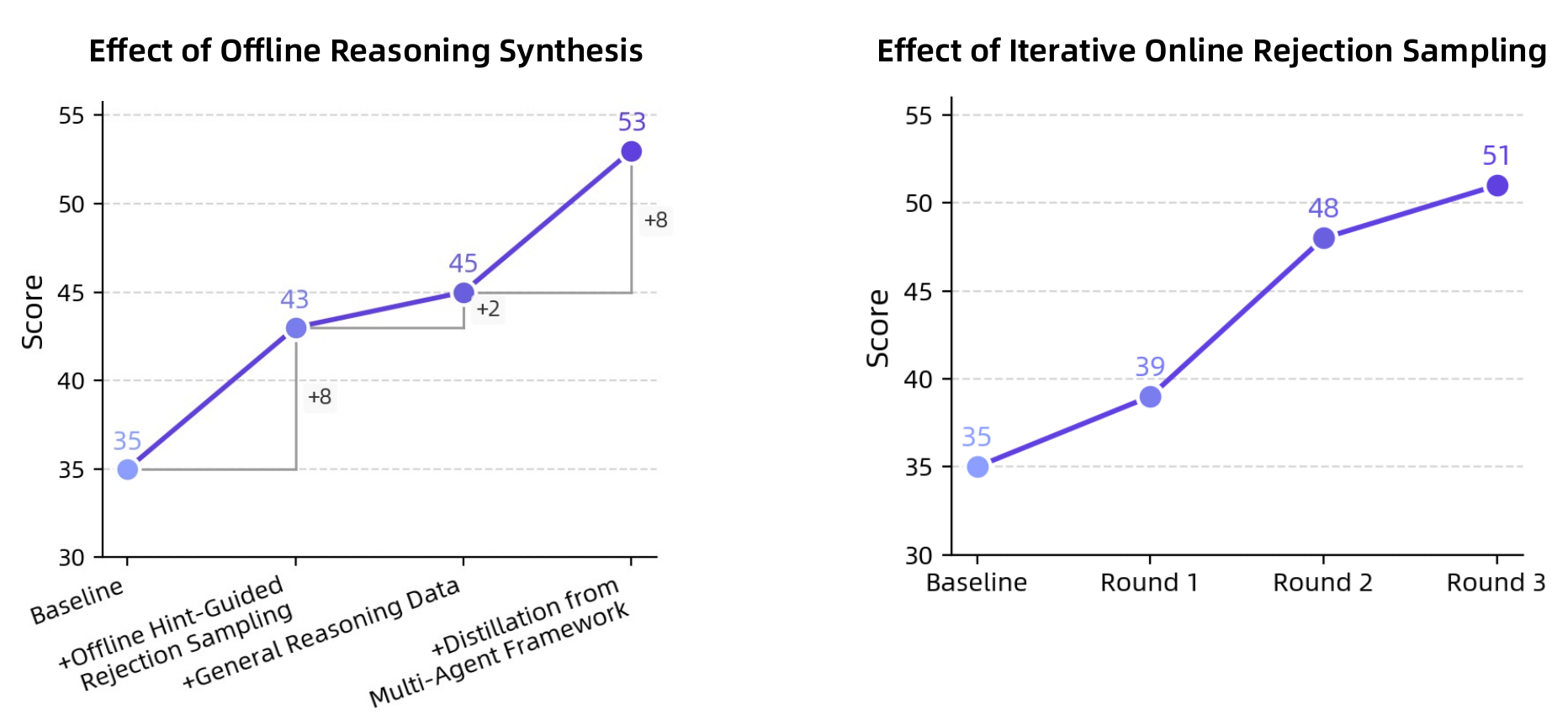}
    \caption{Effect of reasoning data synthesis on Android World.}
    \label{fig:abl_reasoning}
\end{figure}
Our offline reasoning data synthesis primarily comes from two methods: Offline Hint-Guided Rejection Sampling and Distillation from Multi-agent Framework. We also mix in general-purpose reasoning SFT data to maintain the model’s generalization. Beyond the offline data, we use online iterative sampling, continually leveraging updated models to synthesize trajectories with reasoning. We analyze these components separately in \Cref{fig:abl_reasoning}.

We begin validation from an early checkpoint and use performance on AndroidWorld to assess the impact of reasoning synthesis. First, we observe that as we incrementally add data from Offline Hint-Guided Rejection Sampling, distillation from a multi-agent framework, and general-purpose reasoning SFT data, the model’s performance steadily improves. Moreover, adding general reasoning data yields a modest performance gain, indicating that maintaining general reasoning capability is also important for GUI interaction reasoning.

We also examine the gains from iterative training. Starting from the same checkpoint and iteratively training with newly updated trajectory data, we observe sustained performance improvements. It is because, as the model’s reasoning ability improves, an increasing share of tasks in the training query set can be completed, thereby enriching the diversity of the training data and enabling the model to learn more robust reasoning capability.

\subsection{Evaluation on Agentic Frameworks}
\label{sec:agent_adaptability}

To evaluate the adaptability of GUI-Owl in real-world scenarios, we benchmarked its performance as the core vision model within established agentic frameworks. We integrated various VLMs into two distinct setups: the \textbf{Mobile-Agent-E}~\citep{wang2025mobile} framework on the dynamic AndroidWorld environment, and the \textbf{Agent-S2}~\citep{Agent-S2} framework on the OS World desktop environment. This tests the models' ability to generalize across both mobile and PC platforms. The models evaluated include UI-TARS-1.5, UI-TARS-72B, Qwen2.5-VL, Seed-1.5-VL, alongside our GUI-Owl-7B and GUI-Owl-32B.

\begin{table}[t!]
\centering
\label{tab:agent_adaptability}
\resizebox{0.9\columnwidth}{!}{%
\begin{tabular}{l c c}
\toprule
\multirow{2}{*}{\textbf{Model}} & \multicolumn{2}{c}{\textbf{Success Rate (\%)}} \\
\cmidrule(lr){2-3}
& \begin{tabular}[c]{@{}c@{}}\textbf{Mobile-Agent-E~\citep{wang2025mobile}} \\ \textit{on AndroidWorld}\end{tabular} & \begin{tabular}[c]{@{}c@{}}\textbf{Agent-S2~\citep{Agent-S2}} \\ \textit{on a subset of OSWorld-Verified}\end{tabular} \\
\midrule
\multicolumn{3}{l}{\textit{Baseline Models}} \\ \midrule
\quad UI-TARS-1.5~\citep{qin2025ui} & 14.1 & 14.7 \\
\quad UI-TARS-72B~\citep{qin2025ui} & 14.8 & 19.0 \\
\quad Qwen2.5-VL-72B~\citep{Qwen2.5-VL} & 52.6 & 38.6 \\
\quad Seed-1.5-VL~\citep{seed2025seed1_5vl} & 56.0 & 39.7 \\
\midrule
\multicolumn{3}{l}{\textit{Our Models}} \\ \midrule
\quad GUI-Owl-7B  & \underline{59.5} & \underline{40.8} \\
\quad GUI-Owl-32B & \textbf{62.1} & \textbf{48.4} \\
\bottomrule
\end{tabular}%
}
\caption{Performance comparison on agentic frameworks. We report the Success Rate (\%) on both mobile (AndroidWorld) and desktop (OSWorld-Verified) environments. A representative subset of OSWorld-Verified are selected to capture core challenges while reducing computational costs. \underline{Underlined} denotes the second-best performance.}
\end{table}

The experimental results, presented in Table~\Cref{tab:agent_adaptability}, show that GUI-Owl models achieve substantially higher success rates than all baselines on both mobile and desktop platforms. GUI-Owl-32B, in particular, sets the highest result with a score of 62.1 on AndroidWorld and 48.4 on OSWorld. We attribute this superior agentic adaptability primarily to GUI-Owl's enhanced instruction-following capability. Unlike baseline models that may struggle to interpret the specific directives from an agent's planner, GUI-Owl excels at grounding these commands to the correct visual elements on the screen. This leads to more precise action generation (e.g., clicks and text inputs) and critically reduces the accumulation of errors in multi-step tasks. By more reliably executing each step in a sequence, GUI-Owl ensures higher overall task success, making it a more robust and effective "brain" for GUI agents.

\label{section6}
\section{Details of Self-Evolving Trajectory Data Production}

\begin{table}
\begin{tabular}{lp{0.8\textwidth}}
\toprule
\textbf{Action} & \textbf{Definition} \\
\midrule
key & Perform a key event on the mobile device using adb's \texttt{keyevent} syntax. \\ \cmidrule(lr){1-2}
click & Click the point on the screen with specified (x, y) coordinates. \\ \cmidrule(lr){1-2}
long\_press & Press the point on the screen with specified (x, y) coordinates for a specified number of seconds. \\ \cmidrule(lr){1-2}
swipe & Swipe from starting point with specified (x, y) coordinates to endpoint with specified (x2, y2) coordinates. \\ \cmidrule(lr){1-2}
type & Input the specified text into the activated input box. \\ \cmidrule(lr){1-2}
answer & Output the specified answer. \\ \cmidrule(lr){1-2}
system\_button & Press the specified system button: Back, Home, Menu, or Enter. \\ \cmidrule(lr){1-2}
open & Open an application on the device specified by text. \\ \cmidrule(lr){1-2}
wait & Wait for a specified number of seconds for changes to occur. \\ \cmidrule(lr){1-2}
terminate & Terminate the current task and report its completion status: success or failure. \\
\bottomrule
\end{tabular}
\caption{Action Space of \modelname on Mobile.}
\label{tab:space_e2e_mobile}
\end{table}

\begin{table}
\begin{tabular}{lp{0.8\textwidth}}
\toprule
\textbf{Action} & \textbf{Definition} \\
\midrule
key & Performs key down presses on the arguments passed in order, then performs key releases in reverse order. \\
\cmidrule(lr){1-2}
type & Input a string of text. Use the clear parameter to decide whether to overwrite the existing text, and use the enter parameter to decide whether the enter key should be pressed after typing the text. \\
\cmidrule(lr){1-2}
mouse\_move & Move the cursor to a specified (x, y) pixel coordinate on the screen. \\
\cmidrule(lr){1-2}
click & Click the left mouse button at a specified (x, y) pixel coordinate on the screen. \\
\cmidrule(lr){1-2}
drag & Click at a specified (x, y) pixel coordinate on the screen, and drag the cursor to another specified (x2, y2) pixel coordinate on the screen. \\
\cmidrule(lr){1-2}
right\_click & Click the right mouse button at a specified (x, y) pixel coordinate on the screen. \\
\cmidrule(lr){1-2}
middle\_click & Click the middle mouse button at a specified (x, y) pixel coordinate on the screen. \\
\cmidrule(lr){1-2}
double\_click & Double-click the left mouse button at a specified (x, y) pixel coordinate on the screen. \\
\cmidrule(lr){1-2}
scroll & Performs a scroll of the mouse scroll wheel. \\
\cmidrule(lr){1-2}
wait & Wait for a specified number of seconds for changes to occur. \\
\cmidrule(lr){1-2}
terminate & Terminate the current task and report its completion status: success or failure.  \\
\bottomrule
\end{tabular}
\caption{Action Space of \modelname on Desktop.}
\label{tab:space_e2e_desktop}
\end{table}

In this section, we present the details of our self-evolving trajectory data production pipeline.

\subsection{Overview}
GUI automation tasks operate in online interactive environments, which renders manual annotation of trajectory data exceedingly tedious and costly, posing significant challenges for GUI trajectory data collection.
To address these challenges, we develop a self-evolving GUI trajectory data production pipeline. This approach leverages the capabilities of GUI-Owl itself, continuously generating new trajectories through rollout and assessing their correctness to obtain high-quality training data. Subsequently, these data are utilized to enhance the model's capabilities, creating a reinforcing cycle of improvement.

Our pipeline, illustrated in \Cref{fig:data_pipeline},
operates through the following stages:
(1) The process initiates with the construction of online virtual environments encompassing mobile, PC, and web platforms, alongside the generation of diverse queries covering a wide range of potential GUI scenarios;
(2) Given these queries, the GUI-Owl model predicts actions step-by-step, which are then executed within the online virtual environments, yielding roll-out trajectories;
(3) A Trajectory Correctness Judgement module, incorporating a multimodal critic framework, evaluates the correctness of all roll-out trajectories. Successful trajectories are collected to create a rich dataset of interaction sequences that capture temporal dependencies and diverse GUI states;
(4) For challenging queries where the GUI-Owl model struggles to produce successful trajectories despite numerous attempts, we introduce a Query-specific Guidance Generation module. This module synthesizes step-level guidance based on ground-truth trajectories produced by human annotation or other models, facilitating GUI-Owl's handling of difficult tasks and enhancing the efficiency of the entire data generation pipeline;
(5) Finally, all processed data is compiled for reinforcement fine-tuning of GUI-Owl. The model undergoes continuous updates, creating a feedback loop where its ability to generate effective roll-out trajectories improves over time, progressively reducing reliance on manual data collection and achieving self-evolution.

This self-evolving data production pipeline effectively addresses the unique challenges of GUI automation tasks, enabling the creation of robust and versatile GUI intelligent agents capable of handling the complexities of modern graphical user interfaces while continuously improving the efficiency and quality of the data production process itself.

\subsection{High-quality Query Generation}
As highlighted in our overview, generating high-quality queries is a critical component of our self-evolving GUI trajectory data production pipeline. These queries need to cover a wide range of possible user intentions and tasks, reflecting the multifaceted nature of GUI interactions. In this section, we present our innovative approach to query generation for mobile and computer applications, which ensures diversity, realism, and accuracy in the produced queries.

\paragraph{Mobile.}

For mobile applications, we develop a screenshot-action framework that captures the essence of user interactions while maintaining controllability and extensibility.
At the core of our query generation process is a human-annotated directed acyclic graph (DAG) $\mathcal{G} = \langle P, A \rangle$ for each task. Here, $P = \{p_1, \ldots, p_n\}$ represents screenshots (e.g., home, ordering, payment), and $A \subseteq P \times P$ defines valid transitions between them. Each screenshot $p_i$ includes a description $d_i$ of the screenshot's content and purpose and a set of available slot-value pairs that represent possible user choices or inputs on that screenshot. 
This structure allows us to model realistic navigation flows within apps and capture the multi-constraint nature of user queries.

%


Specifically, our query generation process involves the following steps: (1) Path Sampling: We sample a path $P' = \{p_{\sigma_1}, \ldots, p_{\sigma_k}\}$ on the DAG $\mathcal{G}$. This path represents a realistic sequence of screenshot transitions within the app. (2) Metadata Extraction: From the sampled path, we obtain screenshot descriptions $D' = \{d_{\sigma_1}, \ldots, d_{\sigma_k}\}$ and the corresponding slot-value pairs $K', V'$. (3) Instruction Synthesis: The extracted metadata is fed to a Large Language Model (LLM) to synthesize constrained instructions. This approach ensures that the generated queries are both realistic and aligned with the app's structure. (4) Refinement: To enhance naturalness, we refine the raw DAG paths using few-shot LLM prompting. This step transforms explicit navigation instructions into more natural user queries. For example, "Open the takeout app, click on the food entry" becomes "Order me takeout". (5) Interface Validation: To maintain accuracy, we employ web crawlers to collect real-time interface data from target applications. This ensures that aligned with current app functionality.
In conclusion, in our screenshot-action framework, the use of manually defined slot-value pairs minimizes LLM hallucinations, while the DAG structure ensures realistic and controllable navigation flows. 


\paragraph{Computer.}
To acquire operational trajectories for the training of intelligent agents, the initial and most crucial step is the batch acquisition of command data. Unlike mobile phones, the computer usage domain typically involves productivity applications, such as web browsers, document editors, file explorers, and email clients. When it comes to intelligent agents, the utilization of these software tools via keyboard and mouse manipulations presents two primary challenges.

Firstly, there is the fundamental issue of atomic operational skills. Humans, after learning, can proficiently use a mouse for clicking and scrolling and a keyboard for input and shortcut execution. However, intelligent agents driven by vision-language models often lack basic knowledge of atomic operations, such as scrolling through content on web pages or selecting editing targets via dragging in office software.

Secondly, software operational pathways must be navigated, such as accessing privacy settings in Chrome or adjusting page margins in Microsoft Word. Accomplishing these objectives necessitates a series of actions, including clicks, scrolls, and inputs, to reach the requisite configuration options.

Therefore, to bestow intelligent agents with computer usage capabilities, we have synthesized user instructions, targeting both atomic operational skills and software operational pathways, through a combination of manual annotation and automated generation facilitated by Large Language Models (LLMs).

1) Atomic Operations: For common atomic operations with the mouse and keyboard, we initially acquired operational objects within a PC environment via manual annotation. Examples include:
   a) Double-clicking: This involves creating software icons, folders, etc., to train the model in double-click operations.
   b) Input: This involves creating files in formats such as Word, Excel, and PowerPoint to train the model’s capability to accurately input text at specified locations.
   c) Dragging: Similarly, files in Word, Excel, and PowerPoint formats are created to train the model to select specific text or move objects through dragging.

Once operational objects are obtained, we input screenshots of these objects and exemplar commands into the Vision-Language Model (VLM), leveraging its in-context learning capabilities to generate additional executable commands within the current page.

2) Software Operational Pathways: For common software operational pathways, we devised a set of automated deep-search chains. Utilizing an accessibility (a11y) tree, we acquire positional and functional information of actionable elements within software interfaces, and by integrating operational pathway memory and replay, we achieve a tree-structured search of actionable elements (e.g., multi-level menus) to garner corresponding operational pathways.

The endpoint settings of each operational pathway pertain to disparate objects. For example, some configurations alter global file attributes (such as image scaling), whereas others necessitate pre-selecting operational objects (such as altering the font size of a text segment).

Therefore, to derive legally executable commands based on operational paths, we employ an LLM to ascertain whether an operational pathway requires pre-selection of an operational object. For pathways necessitating selected objects, we input manually annotated file screenshots and operational pathways into the VLM, thereby generating commands executable within the current page.



\subsection{Trajectory Correctness Judgment Module}

The Trajectory Correctness Judgment Module plays a crucial role in our self-evolving GUI trajectory data production pipeline. Its primary purposes are twofold: to assess the correctness of roll-out trajectories generated by the GUI-Owl model, and to cleanse erroneous steps within otherwise correct trajectories, thereby enhancing the overall quality of our training data. This module is essential for maintaining high standards in our data collection process, ensuring that only accurate and complete trajectories are used for model training.

Our approach to trajectory correctness judgment is comprehensive, operating at both the step level and the trajectory level. This two-tiered system allows for a nuanced evaluation of each action within a trajectory, as well as an overall assessment of the entire interaction sequence.

\paragraph{Problem Definition of Trajectory Correctness Judgment.}
\label{sec: problem definition}
GUI automation tasks can be formalized as a Markov Decision Process: $\mathcal{M} = (\mathcal{E}, \mathcal{A}, \mathcal{P})$, where $E$ represents the environment state (including user instructions, interaction history, and screenshots), $A$ is the action space, and $P$ is the transition probability.

The Trajectory Correctness Judgement Module consists of two interconnected components: (1) Step-Level Critic: it evaluates individual actions within a trajectory. It analyzes the pre-action state, the executed action, and the post-action state to determine the appropriateness of each step. (2) Trajectory-Level Critic: This component assesses the overall correctness of the entire trajectory. It utilizes the outputs from the Step-Level Critic along with the original user instruction to make a final judgment on the trajectory's success in accomplishing the user's goal.

The relationship between these two levels is hierarchical and complementary. The Step-Level Critic provides granular insights into each action, which are then synthesized by the Trajectory-Level Reflection to form a holistic evaluation of the entire interaction sequence.

\paragraph{Step-Level Critic.}
Achieving a reliable Step-Level Critic presents a sophisticated challenge that demands nuanced environmental perception and comprehensive understanding. The methodology necessitates a meticulous analysis of pre- and post-action screenshots, coupled with a detailed examination of the executed operation, to accurately assess its contribution towards fulfilling the user's designated objective.

Initially, we annotate the critical interaction regions on the pre-action screenshot, enabling the model to focus on pivotal areas of intervention, including precise operational details—such as clicking, long-pressing, or scrolling—to facilitate a comprehensive evaluation of the action's alignment with the user's goal.

Formally, Step-Level Critic can be conceptualized as a function $\pi_{critic}^{step}(\epsilon, a, \epsilon')$, where $\epsilon$ represents the pre-action environmental state (encompassing user instructions, interaction history, and the initial screenshot), $a$ denotes the executed operation, and $\epsilon'$ encapsulates the post-action environmental state. The function generates three critical outputs:

\begin{itemize}
    \item An \textbf{analysis} $a \in \mathcal{A}$ that provides a detailed interpretation of the action's context and consequences
    \item A \textbf{summary} $s \in \mathcal{S}$ that concisely captures the key insights of the action (typically within 30 words)
    \item An \textbf{annotation} $l \in \{GOOD, NEUTRAL, HARMFUL\}$ that categorizes the action's effectiveness towards the user's objective
\end{itemize}

This detailed evaluation at the step level is crucial for identifying and potentially correcting erroneous actions within trajectories, thus improving the overall quality of our training data.


\paragraph{Trajectory-Level Critic.}
The Trajectory-Level Critic, $\pi_{critic}^{traj}(I, T, \pi_{critic}^{step})$, where $I$ represents the user instruction and $T$ represents the action trajectory, provides a comprehensive evaluation of the entire trajectory. It employs a two-channel approach: (1) Textual Reasoning Channel ($\pi_{text}$): Utilizes large language models to assess trajectory correctness based on screenshot caption, textual summaries of each step. (2) Multi-Modal Reasoning Channel ($\pi_{multimodal}$): Incorporates both visual screenshots and textual summaries for a more comprehensive evaluation.
The textual channel provides concise semantic reasoning, while the multi-modal channel enriches the analysis with visual context,
the combination of them helps to mitigate potential biases and limitations inherent in single-modal evaluation.

The final GUI trajectory correctness is determined by a consensus mechanism:

\begin{equation}
    \text{Trajectory Correctness} = \begin{cases} 
    \text{Correct}, & \text{if } \pi_{text}(T, I) = \text{Correct} \land \pi_{multimodal}(T, I) = \text{Correct} \\
    \text{Incorrect}, & \text{otherwise}
    \end{cases}
\end{equation}

In conclusion, this multi-channel approach enhances robustness, processes complementary information, and ensures rigorous validation of trajectories.

\subsection{Query-specific Guidance Generation}
In our constructed query set, some queries pose significant challenges for the model, potentially requiring numerous rollouts to obtain a successful trajectory. Some other queries are even more insurmountable and necessitate manual annotation for reference operational trajectories.
To acquire more diverse training data, we devise a Query-specific Guidance Generation module, which leverages existing successful trajectories to generate guidance that assists the model in producing more successful trajectories.

Initially, for the obtained reference trajectories, we employ a VLM to generate descriptions of the outcomes of each action. Specifically, the input consists of screenshots of the screen before and after the action execution, coupled with the model's or human's action decisions. The VLM is prompted to observe and describe the result of the current action execution, such as "clicked and activated the search box" or "entered the number 100". When actions involve coordinates, we annotate the interaction locations with circles on the pre-action screenshots to help the VLM focus on detailed screen changes.

Regarding the reference trajectories obtained from model rollouts, given the considerable difficulty of the queries, errors or ineffective operations are inevitable. Thus, the VLM also refers to the model's decision rationale, determining whether the outcomes of each step align with the model's expectations. Operations that do not meet expectations or fail to elicit effective responses are subsequently filtered out during the guidance synthesis process.

After acquiring descriptions for each step of the action execution results, we concatenate the descriptions for all steps within the trajectory. Utilizing a LLM, we summarize the essential steps required to complete the query, thereby yielding query-specific guidance.

\subsection{Examples of Training Data}

\begin{figure}[!t]
    \centering
    \includegraphics[width=\textwidth]{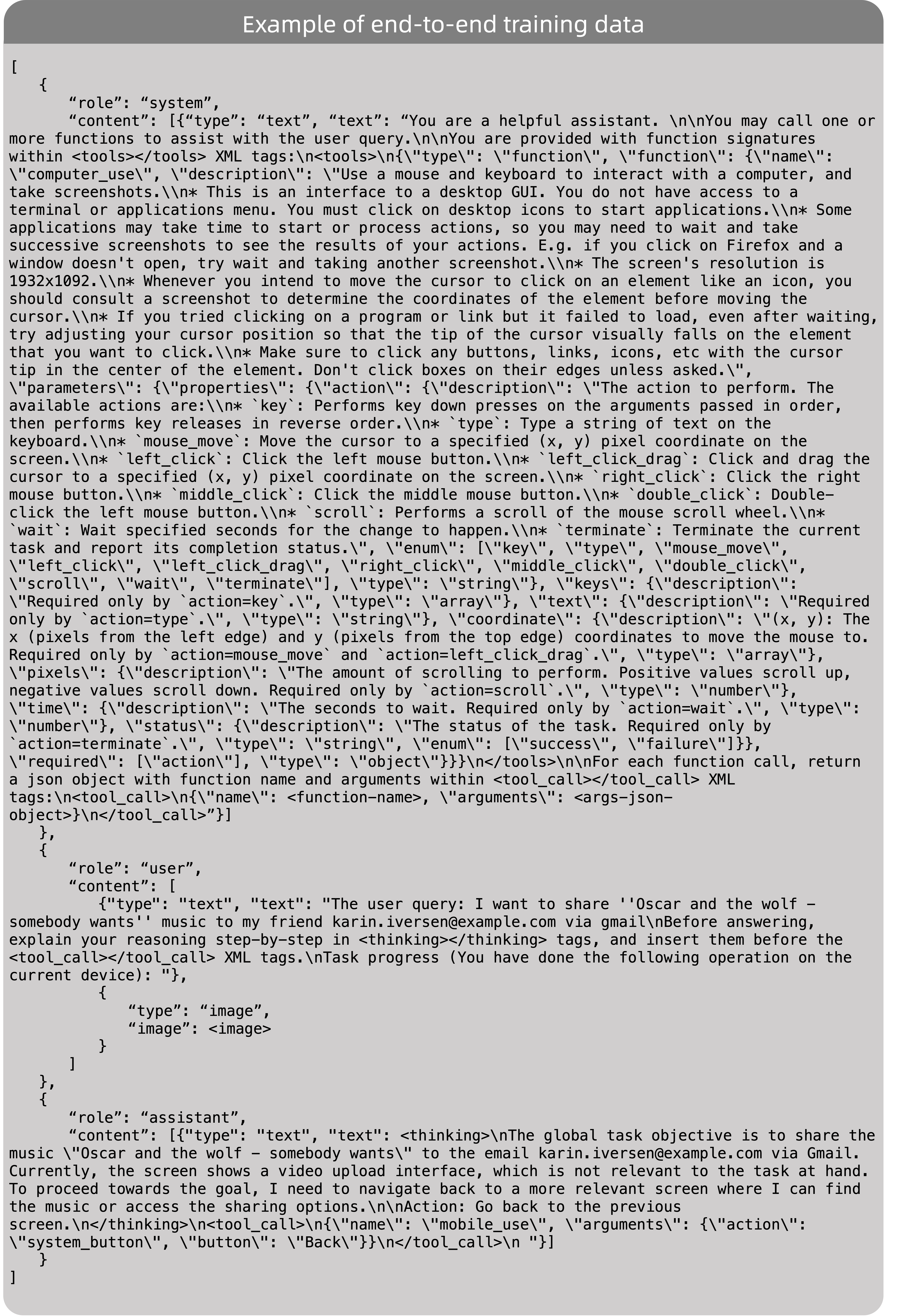}
    \caption{Format of end-to-end training data.}
    \label{fig:format}
\end{figure}

We show the format of end-to-end training data on a desktop platform in Figure~\ref{fig:format}.

\section{Details of Mobile-Agent-v3}
\label{section7}

\subsection{Core Components and Formalism}

The operational dynamics of the Mobile-Agent-v3 framework are defined by a set of state variables and the specialized functions of its constituent agents. We formalize these components as follows.

\subsubsection{State Variables and Definitions}

Let the entire process be a sequence of operations indexed by timestep $t \in \{0, 1, \dots, T\}$.

\begin{itemize}
    \item \textbf{Device State ($S_t$)}: The state of the GUI device at timestep $t$, represented as a high-dimensional tensor $S_t \in \mathcal{S} \subseteq \mathbb{R}^{H \times W \times C}$, where $H, W, C$ are the height, width, and channel dimensions of the screen capture, respectively. $S_0$ denotes the initial state.
    
    \item \textbf{Subsequent Subgoals ($SS_t$)}: An ordered list of pending subgoals formulated by the Manager Agent. It is defined as $CS_t = (g_1, g_2, \dots, g_k)$, where each $g_i$ is a natural language string describing a discrete step towards the main goal.

    \item \textbf{Compltetd Subgoals ($CS_t$)}: A set containing subgoals that have been successfully executed and verified. It is defined as $SS_t = \{\bar{g}_1, \bar{g}_2, \dots, \bar{g}_m\}$. This prevents redundant operations and tracks progress.

    \item \textbf{Action ($A_t$)}: The operation executed by the Worker Agent at timestep $t$. An action is a structured tuple $A_t = (\tau_t, \alpha_t, \sigma_t) \in \mathcal{A}$, where:
    \begin{itemize}
        \item $\tau_t$: The thought process, a textual rationale for selecting the action.
        \item $\alpha_t$: The concrete, low-level action command (e.g., click(x, y), type("text")).
        \item $\sigma_t$: A concise summary of the action's intended effect.
    \end{itemize}

    \item \textbf{Reflection Feedback ($F_t$)}: The output generated by the Reflector Agent after observing the consequences of action $A_t$. It is a tuple $F_t = (j_t, \phi_t) \in \{\text{SUCCESS}, \text{FAILURE}\} \times \Phi$, where:
    \begin{itemize}
        \item $j_t$: A binary judgment on the outcome of $A_t$.
        \item $\phi_t$: A detailed textual feedback, particularly a diagnostic analysis in case of "FAILURE". $\Phi$ represents the space of all possible feedback texts.
    \end{itemize}

    \item \textbf{Notes ($N_t$)}: A collection of critical, potentially transient information captured by the Notetaker Agent. The cumulative knowledge base at step $t$ is $\mathcal{N}_t = \bigcup_{i=0}^{t-1} N_i$.
\end{itemize}

\subsection{Agent Architecture in Detail}

\subsubsection{External Knowledge Retrieval with RAG}
To enable the agent to complete tasks requiring real-time information or domain-specific knowledge (e.g., checking today's weather, finding recent sports scores, or looking up app-specific tutorials), we incorporate a RAG module. This module is invoked at the beginning of a task to retrieve relevant information from external sources, such as the internet, and provide it as context to the agent system.

The process can be formalized as follows. Given an initial user instruction $I$, the RAG module first processes it into one or more search engine-friendly queries $Q$.
\[
Q = \text{GenerateQueries}(I)
\]
Subsequently, the system uses these queries $Q$ to retrieve a set of relevant documents or text snippets $D = \{d_1, d_2, \dots, d_n\}$ from an external knowledge source (e.g., a web search engine).
\[
D = \text{SearchEngine}(Q)
\]
Finally, the retrieved content is processed and summarized to form a concise, information-rich body of knowledge, $K_{\text{RAG}}$.
\[
K_{\text{RAG}} = \text{Process}(D)
\]
This retrieved knowledge $K_{\text{RAG}}$ is passed to the Manager agent during its initialization phase (as shown in Algorithm~\Cref{alg:main_loop}, lines 3-4). This allows the Manager to generate its initial plan ($SS_0, CS_0$) based on more comprehensive and accurate information, thereby significantly improving the quality of the plan and the likelihood of task success. For example, for an instruction like "Should I take an umbrella to the park today?", $K_{\text{RAG}}$ would contain the weather forecast, enabling the Manager to create a plan that includes steps like "open weather app" and "check for rain probability".

\subsubsection{The Manager Agent: Dynamic Task Planning and Coordination}
The Manager Agent serves as the strategic core of the framework. Its function $\mathcal{M}$ is responsible for decomposing the high-level user instruction $I$ into a coherent sequence of subgoals and dynamically adapting this plan throughout the execution process.

Initially, at $t=0$, the Manager performs a decomposition:
\begin{equation}
    (SS_0, CS_0) = \mathcal{M}_{\text{init}}(I, S_0, K_{\text{RAG}})
\end{equation}
where $K_{\text{RAG}}$ is external knowledge retrieved by the RAG module to inform the decomposition of potentially domain-specific or complex instructions. $CS_0$ is initialized as an empty set $\emptyset$.

In subsequent steps $t > 0$, the Manager updates the plan based on the latest execution results:
\begin{equation}
    (SS_t, CS_t) = \mathcal{M}_{\text{update}}(I, S_{t-1}, SS_{t-1}, CS_{t-1}, A_{t-1}, F_{t-1}, \mathcal{N}_t)
\end{equation}
If the previous action was successful ($j_{t-1} = \text{SUCCESS}$), the Manager identifies the completed subgoal in $SS_{t-1}$, moves it to $CS_t$, and re-prioritizes the remaining tasks in $SS_t$. If the action failed ($j_{t-1} = \text{FAILURE}$), the Manager leverages the diagnostic feedback $\phi_{t-1}$ to revise the plan. This may involve re-ordering subgoals, modifying an existing subgoal, inserting a new corrective subgoal, or even reverting to a previous strategy.

\subsubsection{The Worker Agent: Grounded Action Execution}
The Worker Agent is the tactical executor, translating the strategic plan from the Manager into concrete interactions with the GUI. Its function $\mathcal{W}$ aims to execute the highest-priority, currently feasible subgoal from the guidance list $CS_t$.
\begin{equation}
    A_t = \mathcal{W}(I, S_t, SS_t, F_{t-1}, \mathcal{N}_t)
\end{equation}
Upon receiving the subgoal list $SS_t$, the Worker inspects a small subset from the top of the list (e.g., the top $N$ subgoals). It analyzes the current screen $S_t$ to determine which of these subgoals is most relevant and actionable. The decision-making process is informed by feedback from the previous step, $F_{t-1}$, to avoid repeating errors, and the accumulated notes, $\mathcal{N}_t$, to utilize previously stored information (e.g., using a password saved in notes). The output, $A_t = (\tau_t, \alpha_t, \sigma_t)$, provides a transparent record of its reasoning, action, and intent, which is crucial for reflection.

\subsubsection{The Reflector Agent: Self-Correction through Reflection}
The Reflector Agent is a critical component for ensuring robustness and learning from mistakes. It embodies the framework's capacity for self-assessment. Its function $\mathcal{R}$ evaluates the efficacy of an action by comparing the state transition with the Worker's intent.
\begin{equation}
    F_t = \mathcal{R}(I, S_t, S_{t+1}, A_t)
\end{equation}
The Reflector analyzes the pre-action state $S_t$, the post-action state $S_{t+1}$, and the action tuple $A_t$. A judgment $j_t = \text{SUCCESS}$ is rendered if the state change $S_t \rightarrow S_{t+1}$ aligns with the progress articulated in the Worker's thought $\tau_t$ and summary $\sigma_t$. Conversely, $j_t = \text{FAILURE}$ is returned if the GUI presents an error, remains unchanged unexpectedly, or transitions to an irrelevant state. In case of failure, the feedback $\phi_t$ provides a causal analysis, such as action click(123, 456) on button "Submit" did not proceed to the next page; an error message "Invalid credentials" is now visible. This detailed feedback is vital for the Manager's replanning phase.

\subsubsection{The Notetaker Agent: Persistent Contextual Memory}
The Notetaker Agent addresses the challenge of state volatility in GUI interactions, where crucial information may appear on one screen and be required on a subsequent, different screen. The Notetaker's function $\mathcal{C}$ is to identify and persist such information.
\begin{equation}
    N_t = \mathcal{C}(S_t)
\end{equation}
This agent is triggered only upon a successful action ($j_t = \text{SUCCESS}$). It scans the state transition for pieces of information designated as vital for the ongoing task (e.g., reservation codes, order numbers, user-generated content, entered credentials). This information is structured into the note set $N_t$. The cumulative notes $\mathcal{N}_{t+1} = \mathcal{N}_t \cup N_t$ are then made available to both the Manager and the Worker in future steps, creating a persistent memory that informs long-horizon planning and execution.

\begin{algorithm}[!h]
\caption{Mobile-Agent-v3 Execution Loop}
\label{alg:main_loop}
\begin{algorithmic}[1]
    \State \textbf{Input:} User instruction $I$, initial device state $S_0$, max timesteps $T_{\text{max}}$
    \State \textbf{Initialize:} Manager $\agentset{M}$, Worker $\agentset{W}$, Reflector $\agentset{R}$, Notetaker $\agentset{C}$
    \Statex
    \Statex \(\triangleright\) \textbf{\textit{Init Manager Phase: Initialize Plan}}
    \State Retrieve external knowledge $K_{\text{RAG}} \gets \text{RAG}(I)$
    \State $(SS_0, CS_0) \gets \agentset{M}_{\text{init}}(I, S_0, K_{\text{RAG}})$
    \State $\mathcal{N}_0 \gets \emptyset$, $F_{-1} \gets \text{null}$, $t \gets 0$
    \Statex
    \While{$t < T_{\text{max}}$ and $SS_t \neq \emptyset$}
        \Statex\hspace{\algorithmicindent}\(\triangleright\) \textbf{\textit{Worker Phase: Execute Action}}
        \State $A_t \gets \agentset{W}(I, S_t, SS_t, F_{t-1}, \mathcal{N}_t)$
        \If{$A_t = \text{TERMINATE}$}
            \State \textbf{Break}
        \EndIf
        \State $S_{t+1} \gets \text{ExecuteOnDevice}(A_t)$
        \Statex
        \Statex\hspace{\algorithmicindent}\(\triangleright\) \textbf{\textit{Reflector Phase: Evaluate Outcome}}
        \State $F_t \gets \agentset{R}(I, S_t, S_{t+1}, A_t)$
        \Statex
        \Statex\hspace{\algorithmicindent}\(\triangleright\) \textbf{\textit{Notetaker Phase: Persist Information}}
        \If{$F_t.\text{status} = \text{SUCCESS}$}
            \State ${N}_t \gets \agentset{C}(S_t)$
            \State $\mathcal{N}_{t+1} \gets \mathcal{N}_t \cup \{{N}_t\}$
        \Else
            \State $\mathcal{N}_{t+1} \gets \mathcal{N}_t$
        \EndIf
        \Statex
        \Statex\hspace{\algorithmicindent}\(\triangleright\) \textbf{\textit{Manager Phase: Update Plan}}
        \State $(SS_{t+1}, CS_{t+1}) \gets \agentset{M}_{\text{update}}(I, S_t, SS_t, CS_t, A_t, F_t, \mathcal{N}_{t+1})$
        \State $t \gets t + 1$
    \EndWhile
    \Statex
    \If{$SS_t = \emptyset$}
        \State \Return Task Succeeded
    \Else
        \State \Return Task Failed (Timeout or Stalemate)
    \EndIf
\end{algorithmic}
\end{algorithm}
\subsection{Integrated Workflow and Algorithm}

The Mobile-Agent-v3 framework operates as a cyclical, state-driven process. The workflow begins with a user instruction and terminates when the task is complete or deemed unachievable. The entire process is formalized in Algorithm~\ref{alg:main_loop}.

The process is initialized with a high-level user instruction $I$. The Manager Agent, aided by the RAG module, creates an initial subgoal plan $SS_0$. The system then enters an iterative loop. In each iteration $t$, the Worker Agent selects and executes a subgoal, resulting in action $A_t$. The environment transitions to a new state $S_{t+1}$. The Reflector Agent evaluates this transition, producing feedback $F_t$. If the action was successful, the Notetaker Agent may record pertinent information as $N_t$. Finally, the Manager Agent updates the task plan to $(SS_{t+1}, CS_{t+1})$ based on the feedback.

Termination occurs under two conditions:
\begin{enumerate}
    \item \textbf{Task Completion}: The Manager determines the task is complete, resulting in an empty pending subgoal list ($SS_t = \emptyset$).
    \item \textbf{Execution Stalemate}: The Worker Agent determines that no pending subgoals in $SS_t$ can be executed on the current state $S_t$, even after several retries or plan revisions.
\end{enumerate}

This structured, reflective, and adaptive loop enables the framework to navigate complex sequences of interactions, handle unexpected events, and robustly pursue the completion of the user's goal.

\subsection{Case Study}

\begin{figure}[!t]
    \centering
    \includegraphics[width=\textwidth]{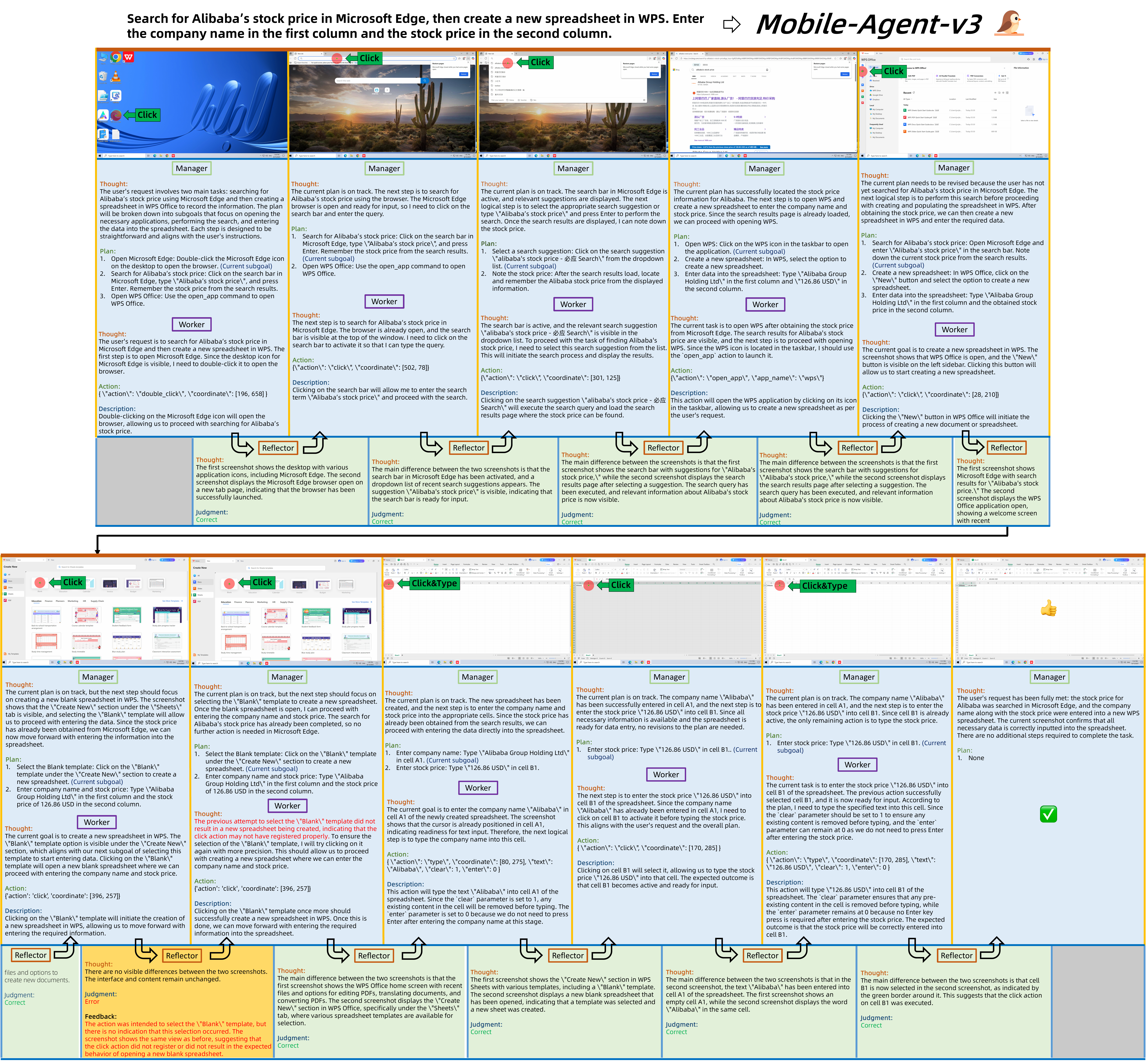}
    \caption{A case of a complete Mobile-Agent-3 operation process on a desktop platform. The red text represents successful reflection content.}
    \label{fig:case}
\end{figure}

Figure~\ref{fig:case} shows a complete Mobile-Agent-v3 operation flow, including the outputs of the manager, worker, and reflector. The manager's output shows that subgoals are constantly updated as the task progresses. The worker consistently outputs actions guided by the subgoals output by the manager. Notably, the red text in Figure~\ref{fig:case} illustrates a successful reflection. After the worker's click operation in the previous step failed, the reflector successfully discovered the problem and provided feedback to the manager and worker in the next step. Finally, the worker repeated the click operation to correct the problem.

\section{Conclusion}
In this paper, we present GUI-Owl, a native end-to-end multimodal agent model that unifies perception, grounding, reasoning, planning, and action execution within a single scalable framework for GUI automation. Building upon Qwen2.5-VL and extensively post-trained on large-scale, diverse GUI interaction data, GUI-Owl achieves state-of-the-art performance across a broad range of challenging benchmarks, surpassing both open-source and proprietary systems, including GPT-4o and Claude 3.7. Through synthesized reasoning data and a scalable reinforcement learning framework, GUI-Owl is capable of versatile decision-making from autonomous single-agent execution to collaborative multi-agent role coordination within our Mobile-Agent-v3 framework.

\clearpage
\bibliography{iclr2024_conference}
\bibliographystyle{iclr2024_conference}

\clearpage
\appendix

\end{document}